\documentclass{article} % For LaTeX2e
\usepackage[preprint]{colm2026_conference}

\usepackage{microtype}
\usepackage{hyperref}
\usepackage{url}
\usepackage{booktabs}

\usepackage{graphicx}
\usepackage{amsmath}
\usepackage{amssymb}
\usepackage{tabularx} 
\usepackage{xspace}
\usepackage{multirow}
\usepackage{float}
\usepackage{caption}
\definecolor{mygray}{RGB}{169,169,169}
\definecolor{myblue}{RGB}{0,102,204}
\definecolor{mygreen}{RGB}{0, 128, 0}
\definecolor{myred}{RGB}{255, 0, 0} 
\usepackage{colortbl}
\usepackage{url}
\usepackage{hyperref}
\usepackage{mathrsfs}
\usepackage{enumitem}
\usepackage{wrapfig}
\usepackage{xcolor}
\usepackage{mdframed}

\usepackage{tcolorbox}
\usepackage{natbib}
\tcbuselibrary{skins,breakable}
\newcommand\cggender{{GCGender}\xspace}
\newcommand\InclusiveGender{{InclusiveGender}\xspace}

\usepackage{lineno}

\definecolor{darkblue}{rgb}{0, 0, 0.5}
\hypersetup{colorlinks=true, citecolor=darkblue, linkcolor=darkblue, urlcolor=darkblue}

\title{Neuron-Level Interventions for Gendered and Gender-Neutral Generation in Language Models}

\author{
  \textbf{Zhiwen You}\textsuperscript{1}\thanks{email: \texttt{zhiweny2@illinois.edu}} \quad
  \textbf{Nafiseh Nikeghbal}\textsuperscript{2,3} \quad
  \textbf{Jana Diesner}\textsuperscript{1,2,3} \\[0.7ex]
  \textsuperscript{1}\,University of Illinois Urbana-Champaign \quad
  \textsuperscript{2}\,Technical University of Munich \\
  \textsuperscript{3} Munich Center for Machine Learning 
}

\newcommand{\github}{\raisebox{-1.5pt}{\includegraphics[height=1.00em]{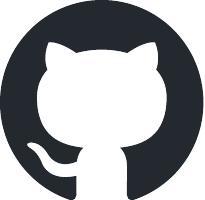}}\xspace}

\begin{document}

\ifcolmsubmission
\linenumbers
\fi

\maketitle

\begin{abstract}
Language models (LMs) can produce gendered language and stereotypes even when given neutral prompts. Most prior work on gender bias in LMs primarily examines gender through a binary lens (feminine vs. masculine), with limited attention to gender-neutral forms, such as they/them pronouns or neutrally phrased job titles. How gender-related signals are encoded in the internal representations of LMs remains an open question. In this work, we study gender-specific neurons in LMs across three categories: feminine, masculine, and gender-neutral. We propose a neuron-level intervention method to identify neurons that are strongly tied to each gender category. We then test these neurons through controlled generation, showing that activating or masking gender-related neurons can steer a sentence toward a target gender form while preserving its original meaning. To evaluate the effectiveness of our gender-intervention approach, we curate two datasets with controlled sentences labeled across all three gender categories and validate the data quality through human evaluation. Experiments on two open-source LMs show that gender-specific neurons are not evenly distributed across model layers; instead, they concentrate heavily in the earliest layers with smaller contributions from later layers. Compared to existing methods, our method achieves more precise gender control, with less leakage into non-target gender categories and stable output quality through two evaluation criteria. Overall, our work examines how gender is encoded in LMs and provides a simple yet effective approach toward controlled gender intervention for both neuron intervention evaluation and gender bias mitigation. Code and datasets are available at: \github\url{https://github.com/zhiwenyou103/Gender-Neuron-Intervention}.

\end{abstract}

\section{Introduction}
Language models (LMs) may encode and generate biased language, including gender stereotypes and unequal associations between gender and occupations~\citep{kotek2023gender,dong2024disclosure,an2025mutual, nikeghbal-etal-2025-cobia}. This is a real-world problem in user-facing settings: a neutral input can still trigger gendered wording, and small wording differences can change who is described as competent, caring, or authoritative. 
Previous studies have investigated the bias issues of LMs in different aspects. Some methods edit LM's outputs with data interventions or structured constraints \citep{thakur2023language,ma2024debiasing,oba2024contextual,you2024sciprompt}. Other work aims to find where a behavior is encoded inside the LM to intervene the bias more directly \citep{liu2024devil,xu2025biasedit,limisiewiczdebiasing}. 

In Transformer-based LMs \citep{vaswani2017attention}, the feed-forward network (FFN) layers contain neurons, and prior work has shown that specific behaviors can be localized to subsets of these neurons \citep{tang2024language, lai2024style}. For example, language-specific neurons can be detected by measuring how often each neuron activates for each language, and then low-entropy neurons can be used to steer the model's output language \citep{tang2024language}. Similarly, style-specific neurons can be found and then deactivated to improve style transfer, though this may affect fluency and requires careful decoding \citep{lai2024style}. Other studies also explore neurons tied to factual relations and shows that deactivating them changes relational recall \citep{liu2025relation}. These findings motivate a question for gender bias study: \emph{Can we find and control gender-related neurons in the same way?}

Most gender bias studies on LMs mainly focus on binary genders (feminine vs.\ masculine). However, in real-world cases, \emph{gender-neutral} words also exist in LM's generation, such as singular ``they,'' neutral role nouns (e.g., \emph{fisher}), and inclusive rewrites that avoid gender-marked terms. In this study, we propose a neuron-level study of gender bias with three gender categories: \emph{feminine}, \emph{masculine}, and \emph{neutral}. To quantify the gender neuron identification quality, we evaluate the performance of transferring gendered sentences: given an input sentence, whether the LM can transfer it into a targeted form (feminine/masculine/neutral) after we deactivating the identified gender neurons, while preserving meaning.
Our contributions include:
\begin{itemize}
  \item We introduce a novel neuron-intervention approach for identifying feminine, masculine, and gender-neutral representations, extending prior binary gender analyses.
  \item We propose a new evaluation protocol to measure the effectiveness of the identified gender neurons.
  \item We curate a new dataset, \InclusiveGender, with 8,600 sentences for each gender category, and expand an existing binary gendered dataset by adding gender-neutral sentences for ternary gender analysis.
\end{itemize}

\section{Related Work}

\subsection{Gender Bias and Stereotypes in LMs}
LMs can produce gendered wording, even when prompts are neutral or underspecified \citep{kotek2023gender,dong2023probing,you2024beyond,lee2025revisiting}. Beyond pronouns, prior work reports associations between gender and social roles, such as occupations, and analyzes how these associations appear in model outputs and representations \citep{an2025mutual}. Other work discusses how to evaluate, and mitigate gender bias in LMs, including guidance on responsible disclosure and practical mitigation choices \citep{dong2024disclosure}.
Several mitigation approaches operate at the inference level, without explicitly exploring the encoded bias inside the LMs internal representations. For example, some studies mitigate gender bias using few-shot data interventions \citep{thakur2023language}, structured knowledge constraints \citep{ma2024debiasing}, or in-context strategies that suppress biased generations at inference time \citep{oba2024contextual}. These methods are effective, but they often provide limited insight into \emph{where} gendered behavior is implemented inside the model. Also, most studies still focus on a binary framing of gender, while gender-neutral forms (e.g., singular \emph{they}, neutral job titles) are less explored, even though they are increasingly recognized as a viable path toward inclusive language generation and translation \citep{piergentili-etal-2023-gender, dawkins-etal-2025-gender, savoldi-etal-2025-mind}.
\subsection{Probing and Controlling Gender Bias}
Previous work probes and intervenes on internal representations to study and control social bias, examining hidden states, attention patterns, or feed-forward activations, and testing causality by modifying these components~\citep{liu2024devil, manna-etal-2025-paying,hackenbuchner2026triggersmodelcontrastiveexplanations, attanasio-etal-2023-tale}. Recent study also proposes targeted interventions such as removing or suppressing bias-related neurons during inference \citep{yang2024mitigating}, deactivating coupled neurons to address fairness-related trade-offs \citep{qian2024dean}, or editing model behavior through model editing techniques \citep{xu2025biasedit, lutz-etal-2024-local}.
Our work is inspired by attribute-specific neuron studies \citep{tang2024language, liu2025relation} that (1) identify a small set of feed-forward neurons linked to an attribute, and (2) steer generation by activating or deactivating those neurons. This approach has been used for multilingual control (language-specific neurons, either natural or programming)~\citep{tang2024language, kojima-etal-2024-multilingual, kargaran-etal-2025-programming, wang2024sharing, stanczak-etal-2022-neurons, zhang-etal-2025-multilingual} and for controlling writing style (style-specific neurons) \citep{lai2024style}. Other work studies neurons tied to factual relations and uses neuron-level interventions to change relational behavior \citep{liu2025relation}. Additionally, representation-space steering methods extract directions (vectors) for attributes and apply them to influence generation \citep{cyberey2025sensing}. 
Compared with prior gender work that mainly probes binary gender or focuses on output-only mitigation, our approach investigates feminine, masculine, and gender-neutral patterns and evaluates neuron interventions through a gender transfer test: whether internal edits causally change gendered wording while preserving meaning.

\section{Method}
\label{sec:method}
\begin{wrapfigure}{r}{0.58\textwidth}
  \vspace{-10pt}
  \centering
  \includegraphics[width=0.58\textwidth]{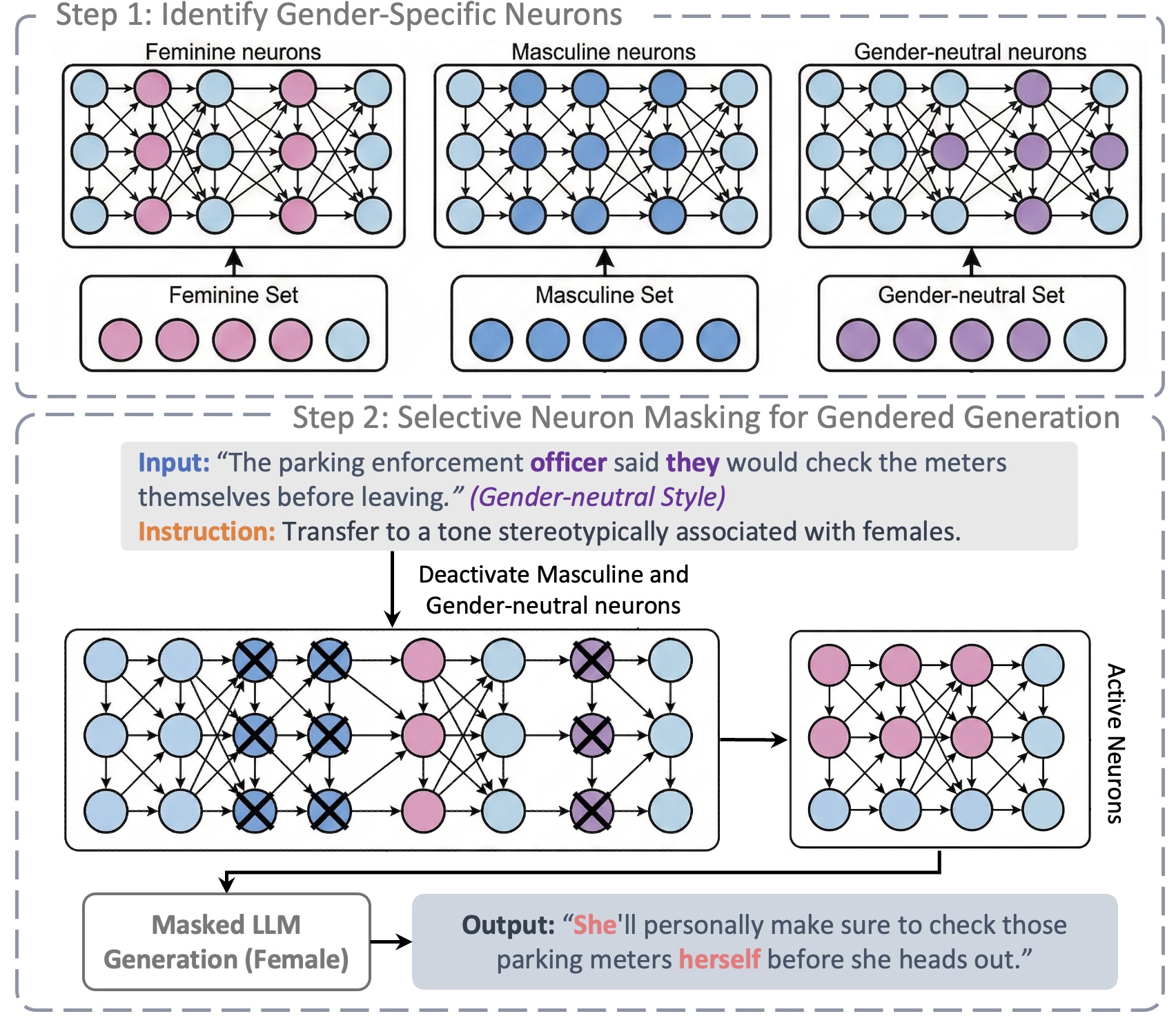}
  \caption{Overview of our gender-specific neuron intervention approach.
  We first identify feminine, masculine, and gender-neutral neurons in the LM. We then selectively mask non-target gender neurons to steer generation toward a target gender, enabling controlled gendered generation while preserving the original semantic content (details in Section~\ref{sec:eval}).
  }
  \label{fig:framework}
  \vspace{-40pt}
\end{wrapfigure}
Here, we introduce our method for intervening gender-related neurons in LMs (Figure~\ref{fig:framework}). We explain our method in the following subsections.
\subsection{Neuron Activation}
We consider three gender categories: Masculine ($m$), Feminine ($f$), and Gender-neutral ($n$). For each dataset, $s_i$ is an input sentence and $g \in \{m, f, n\}$ is its associated gender label. We tokenize each sentence using the model's tokenizer and group them by their respective gender labels to create three distinct subsets for activation analysis.
We focus on the intermediate neurons within the Multi-Layer Perceptron (MLP) blocks of the model. For a model with $L$ layers, $h_j^{(l)}(x)$ represents the activation value of the $j$-th neuron in the $l$-th layer for a given input token $x$ after the activation function (e.g., SiLU \citep{elfwing2018sigmoid}). To investigate how neurons respond to different genders, we calculate the activation level for each neuron across all tokens in a gender-specific subset. For Llama-style gated MLPs, we define the intermediate activations as:
\begin{equation*}
a^{(l)} = \text{SiLU}(W_{\text{gate}}^{(l)} h^{(l-1)}) \odot (W_{\text{up}}^{(l)} h^{(l-1)}),
\end{equation*}
where $h^{(l-1)}$ represents the hidden state from the previous layer, $W_{\text{gate}}^{(l)}$ and $W_{\text{up}}^{(l)}$ are learnable weight matrices, $\text{SiLU}(\cdot)$ is the Sigmoid Linear Unit activation function, and $\odot$ denotes element-wise multiplication.
For each gender category $g \in \{m, f, n\}$, we process the corresponding text corpus through the model and accumulate token-level statistics for every neuron $j$ in layer $l$. The $j$-th neuron of the layer is considered to be active when its accumulated activation value $\bar{a}^{(l)} > 0$.

\subsection{Gender-Specific Neuron Identification and Filtering} \label{sec:identification}

Building on recent work in neuron analysis, we identify neurons that exhibit strong gender-specific behavior using a combined exclusivity scoring approach. Unlike previous methods that focus on language-specific \citep{tang2024language} or generation style neurons \citep{lai2024style}, we address the challenge of overlapping activations across gender categories, which is particularly important given that gender-related features are more subtle than language or style differences.

\textbf{Combined Exclusivity Score.}
For each neuron $j$ in layer $l$, we compute a one-vs-rest exclusivity score for each gender $g$ by comparing its activation statistics against the aggregate statistics of all other genders. We calculate three complementary measures: 
\begin{equation*}
\begin{minipage}{0.30\linewidth} 
\centering
(1) $d_g^{(l,j)} = \frac{\mu_g^{(l,j)} - \mu_{\neg g}^{(l,j)}}{\sqrt{(\sigma_g^{2(l,j)} + \sigma_{\neg g}^{2(l,j)}) / 2}}$, 
\end{minipage}
\begin{minipage}{0.48\linewidth}
\centering
(2) $\Delta_g^{(l,j)} = \log\left(\frac{p_g^{(l,j)}}{1 - p_g^{(l,j)}}\right) - \log\left(\frac{p_{\neg g}^{(l,j)}}{1 - p_{\neg g}^{(l,j)}}\right)$, \end{minipage}
\begin{minipage}{0.22\linewidth}
\centering
(3) $r_g^{(l,j)} = \frac{\mu_g^{(l,j)} - \mu_{\neg g}^{(l,j)}}{|\mu_{\neg g}^{(l,j)}| + \epsilon}$ 
\end{minipage} 
\end{equation*}
(1) \textbf{Effect Size (Cohen's $d$):} Quantifies the standardized difference in mean activations between the target gender and others,
% \begin{equation*}
% d_g^{(l,j)} = \frac{\mu_g^{(l,j)} - \mu_{\neg g}^{(l,j)}}{\sqrt{(\sigma_g^{2(l,j)} + \sigma_{\neg g}^{2(l,j)}) / 2}}
% \end{equation*}
where $\mu_{\neg g}$ and $\sigma_{\neg g}^{2}$ represent the pooled mean and variance across all genders except $g$.

(2) \textbf{Log-Odds Difference ($\Delta$):} Measures the relative likelihood of positive activation for the target gender.
% \begin{equation*}
% \Delta_g^{(l,j)} = \log\left(\frac{p_g^{(l,j)}}{1 - p_g^{(l,j)}}\right) - \log\left(\frac{p_{\neg g}^{(l,j)}}{1 - p_{\neg g}^{(l,j)}}\right).
% \end{equation*}

(3) \textbf{Relative Mean Difference ($r$):} Captures the proportional difference in activation magnitude,
% \begin{equation*}
% r_g^{(l,j)} = \frac{\mu_g^{(l,j)} - \mu_{\neg g}^{(l,j)}}{|\mu_{\neg g}^{(l,j)}| + \epsilon},
% \end{equation*}
where $\epsilon$ is a small constant for numerical stability.

Finally, we normalize each component across all neurons for each $g$, and produce a unified exclusivity score. In neuron selection, a neuron is selected exclusively to gender $g$ if: (1) $g$ has the highest exclusivity score among all genders and (2) the score reaches the pre-defined threshold across multiple criteria (see Appendix~\ref{appx:ablation} for details). This ensures that selected neurons demonstrate clear, unambiguous preference for a single gender category.
\subsection{Masking Gender Neurons for Controlled Generation}
To verify if the identified neurons actually control the model's generation, we perform an intervention during the inference process. We apply a mask to the activations of the identified neurons and generate both \textbf{``baseline''} outputs (no masking) and \textbf{masked} outputs (keep-only masking) for every input sentence. For each input sentence, we apply a prompt for each target gender in masculine, feminine, and gender-neutral. We ask the model to rewrite the input into the target gender form while keeping the meaning unchanged.

\textbf{Baseline Generation.}
We first run the model without any intervention to generate a baseline rewritten sentence for each target gender prompt. All hyper-parameter settings (e.g., temperature, top-$p$, max token length) are kept fixed for fair comparison between baseline and masked runs (see Section~\ref{sec:implementation} for more details).

\textbf{Keep-Only Masked Generation.}
To test whether the identified neurons causally control gender form, we run three masked conditions, one per ``kept'' gender. In this setting, we mask all neurons that belong to the other gender sets, while leaving the kept-gender neurons active. We implement masking by temporarily overriding the MLP forward function in each layer during decoding. For each keep-gender condition, we (1) attach the layer-wise masks, (2) generate outputs for all prompts, and (3) restore the original forward functions to return the model to its unmodified state before the next condition.

\begin{table}[t]
\centering
\footnotesize
\resizebox{0.7\textwidth}{!}{%
\setlength{\tabcolsep}{6pt}
\begin{tabular}{@{}lcc@{}}
\toprule
 & \cggender & \InclusiveGender \\
\midrule
\# of Sent.             & 3,341 / 418 / 417  & 20,640 / 2,580 / 2,580 \\
\# of Sent.\ per Gender & 1,114 / 139 / 139  & 6,880 / 860 / 860 \\
Average Length          & 12 / 12 / 12       & 19 / 20 / 20 \\
\bottomrule
\end{tabular}
}
\caption{Statistics of datasets. We randomly split each dataset into training\allowbreak/validation\allowbreak/testing sets and keep the number of sentences of each gender category the same. The average length is calculated in the token level.}
\label{tab:dataset_stats}
\end{table} 

\section{Datasets} \label{sec:datasets}

We construct two datasets extending prior binary-gender settings with gender-neutral labels: \cggender, by generating neutral sentences based on an existing sentence-level gendered dataset, and \InclusiveGender, by prompting an LM with curated gendered term dictionary and their neutral equivalents (see Table~\ref{tab:dataset_stats} for statistics).

\subsection{\cggender}
We begin with the dataset introduced by \cite{soundararajan2023using}, which consists of synthetically generated English sentences containing gendered language, produced using ChatGPT. Sentence generation is guided by gender-coded lexicons, including the \cite{gaucher2011evidence} lexicon and a larger adjective-based lexicon proposed by \cite{cryan2020detecting}. Each instance is a sentence about a person that includes one or more gender-coded adjectives, targeting stereotypical traits associated with a gender. Sentences are labeled as consistent or contradictory with gender stereotypes depending on whether the gender of the person matches the gender implied by the word. The labels are validated through human annotation and classification experiments, showing strong agreement with human judgments.

Our source dataset \citep{soundararajan2023using} does not include ``gender-neutral'' sentences, and our analysis requires sentences for three gender categories: masculine, feminine, and neutral. To address this, we generate gender-neutral versions of the existing gendered sentences using \texttt{Llama-3.3-70B} served by Ollama~\citep{grattafiori2024llama3herdmodels, llama33_70b_ollama}, replacing gendered terms while preserving sentence meaning. We test multiple prompt variations on a small manually annotated subset of 100 sentences to check accuracy and meaning preservation. The best-performing prompt achieves a binary accuracy of 96\% with most errors being minor over-neutralizations rather than missed gender markers. Then the selected prompt is used to generate gender-neutral sentences for the entire corpus. We combine the Gaucher \citep{gaucher2011evidence} and Cryan \citep{cryan2020detecting} datasets to create a new dataset, \cggender, with 4,176 sentences, including both original gendered and newly generated gender-neutral versions.

\subsection{\InclusiveGender}
In the second approach, we further curate a larger dataset \InclusiveGender by generating sentences for three gender categories: masculine, feminine, and neutral, using the \textit{Inclusive Language} repository \citep{henderson_inclusive_language}, which provides curated lists of gendered terms and their gender-neutral equivalents. This resource serves as the basis for constructing a dataset of sentences representing each gender category. Using these term lists, we design a structured prompt for \texttt{gpt-4o-2024-08-06}~\citep{gpt4o, openai2024gpt4ocard} to generate sentences according to specific rules. Each sentence contains exactly one gender term and at least one corresponding pronoun. The inclusion of pronouns ensures unambiguous category assignment for terms with overlapping usage, such as \textit{actor}, which can appear in both masculine and gender-neutral contexts, in contrast to \textit{actress}. All generated sentences are mutually exclusive and labeled as feminine, masculine, or gender-neutral. The prompts used to generate \InclusiveGender and \cggender are provided in Appendix~\ref{appx:dataset-prompts}.

\begin{table}[t]
\centering
\footnotesize
\resizebox{0.85\textwidth}{!}{%
\setlength{\tabcolsep}{4pt}
\begin{tabular}{lcccccc}
\toprule
 & \multicolumn{3}{c}{\InclusiveGender} & \multicolumn{3}{c}{\cggender} \\
\cmidrule(lr){2-4} \cmidrule(lr){5-7}
Criterion & Ann. 1 & Ann 2 & Agr.($\uparrow$)  & Ann. 1 & Ann. 2 & Agr.($\uparrow$)  \\
\midrule
Gender Correctness ($\uparrow$) & 99.5 & 100.0 & 99.5 & 95.5 & 97.0 & 96.5 \\
Gender Ambiguity ($\downarrow$) & 1.0  & 2.0   & 97.0 & 3.0  & 2.5  & 96.5 \\
Grammatical Correctness ($\uparrow$)     & 96.5 & 99.0  & 96.5 & 98.5 & 97.5 & 96.0 \\
\bottomrule
\end{tabular}
}
\caption{Human annotation results for \InclusiveGender and \cggender 
($N=200$ per dataset). Annotator~1 and Annotator~2 report the percentage of “Yes” annotations for each criterion; Agr. indicates pairwise agreement. $\uparrow$ indicates higher is better; $\downarrow$ indicates lower is better.}
\label{tab:human-annotation}
\end{table}

\subsection{Human Annotation for Dataset Validation}

We validate \cggender and \InclusiveGender via human annotation. Two annotators independently assessed 200 randomly sampled sentences per dataset, stratified across feminine, masculine, and neutral categories.
Each sentence was evaluated along three binary dimensions: (1) gender correctness, indicating whether the sentence matches its assigned gender label; (2) gender ambiguity, indicating the presence of conflicting gender markers; and (3) grammatical correctness, indicating whether the sentence is well-formed in English. As shown in Table~\ref{tab:human-annotation}, both datasets exhibit high annotation quality: gender correctness exceeds 95\%, ambiguity remains below 3\%, and pairwise agreement is above 96\% across all dimensions. For annotation details, see Appendix~\ref{appx:human_instruction}.

\section{Experiments} 
We provide implementation details and the evaluation protocol for our experiments using the datasets introduced in Section~\ref{sec:datasets}.

\subsection{Competing Methods, and Implementation}\label{sec:implementation}

\textbf{Competing Methods.} Following previous studies on neuron identification \citep{lai2024style}, we adopt two existing methods in our study: (1) LAPE: we use the activation probability entropy to identify gender-specific neurons by computing the activation likelihood of individual neurons and selecting neurons with lower entropy scores \citep{tang2024language}; (2) sNeuron-TST: we identify gender-specific neurons by selecting neurons with top activation values for each gender category and eliminating overlapping neurons between source and target genders to avoid ambiguous feature encoding \citep{lai2024style}.

\textbf{Implementation.} We evaluate our proposed gender neuron identification method on two open-source LMs: \texttt{Llama-3.1-8B-Instruct}~\citep{llama31_8b_instruct, grattafiori2024llama3herdmodels} (hereafter \texttt{Llama 3.1}) and \texttt{Qwen2.5-7B}~\citep{qwen2_5_7b, qwen2025qwen25technicalreport} (hereafter \texttt{Qwen 2.5}). For consistent output, we set the temperature of both models as 0 and an maximum output length as 64 tokens. The prompt templates used for gender transfer are provided in Appendix~\ref{appx:prompts}.

\subsection{Evaluation}\label{sec:eval}
We evaluate the effectiveness of gender-specific neuron masking along two dimensions: automatic keyword-based analysis and human judgment.

\textbf{Keyword-Based Gender Term Analysis.} We evaluate masking effectiveness by analyzing gendered terms in generated responses. We curate a dictionary of gender terms associated with three categories: masculine-associated terms (e.g., ``he,'' ``his,'' ``man''), feminine-associated terms (e.g., ``she,'' ``her,'' ``woman''), and neutral terms (e.g., ``they,'' ``person,'' ``individual''). For each generated text, we compute the Term Ratio as the proportion of gender-specific terms relative to total words:
\begin{equation*}
    R_g = \frac{c_g}{w_{\text{total}}},
\end{equation*}
where ${c_g}$ is the count of gender-$g$ terms and $w_{\text{total}}$ is the total word count. We also report the average term count per response across masking conditions. Our curated gendered term lexicon is reported in Appendix~\ref{app:gender_terms}.

We evaluate our approach with a \textit{matching masking} test: for each target gender $g \in \{m, f, n\}$, we compare baseline (unmasked) generation and Keep-only-$g$ masking, where only the neurons identified for $g$ remain active. We report $\Delta$ Ratio (\%) as the change in mention ratio relative to the baseline under the same target. We hypothesize the neuron identification is effective if the target $g$ ratio stays similar or increases ($\Delta g \ge 0$), while the other two $g'$ ratios decrease ($\Delta {g'} < 0$ for $g' \neq g$). This directly tests whether the identified gender neurons causally support the intended gender form while suppressing non-target gender markers.

\textbf{Human Evaluation.} To assess the quality of gender transfer beyond surface-level keyword shifts, we conduct a human annotation study. Annotators evaluate two dimensions: \textit{Overall Idea Preservation}, measuring whether the output retains the core meaning of the input, and \textit{Target Gender Realization}, measuring whether the output correctly reflects the intended target gender. We provide the results in Section~\ref{sec:results} and human
annotation process and the instructions provided to annotators in Appendix~\ref{appx:human_instruction}.

\section{Results and Analysis} \label{sec:results}

We conduct experiments using \texttt{Qwen 2.5} and \texttt{Llama 3.1} on the \InclusiveGender and \cggender test sets, first analyzing layer-by-layer gender neuron distributions (Section~\ref{sec:neuron-distribution}), then reporting quantitative evaluation results (Section~\ref{sec:main-results}), and finally assessing gender transformation quality through human evaluation (Section~\ref{quality-eval}).

% \begin{figure*}[ht]
%   \centering
%   \includegraphics[width=0.7\textwidth]{pics/layer_wise_distribution.png}
%   \caption{Statistics of the number of gender-specific neurons using our method in each layer in \texttt{Llama 3.1} and \texttt{Qwen 2.5} on \InclusiveGender and \cggender datasets.}
%   \label{fig:neuron-distribution}
% \end{figure*}

\subsection{Gender Neuron Distribution} \label{sec:neuron-distribution}
\begin{wrapfigure}{r}{0.6\textwidth}
  \vspace{-15pt}
  \centering
  \includegraphics[width=0.6\textwidth]{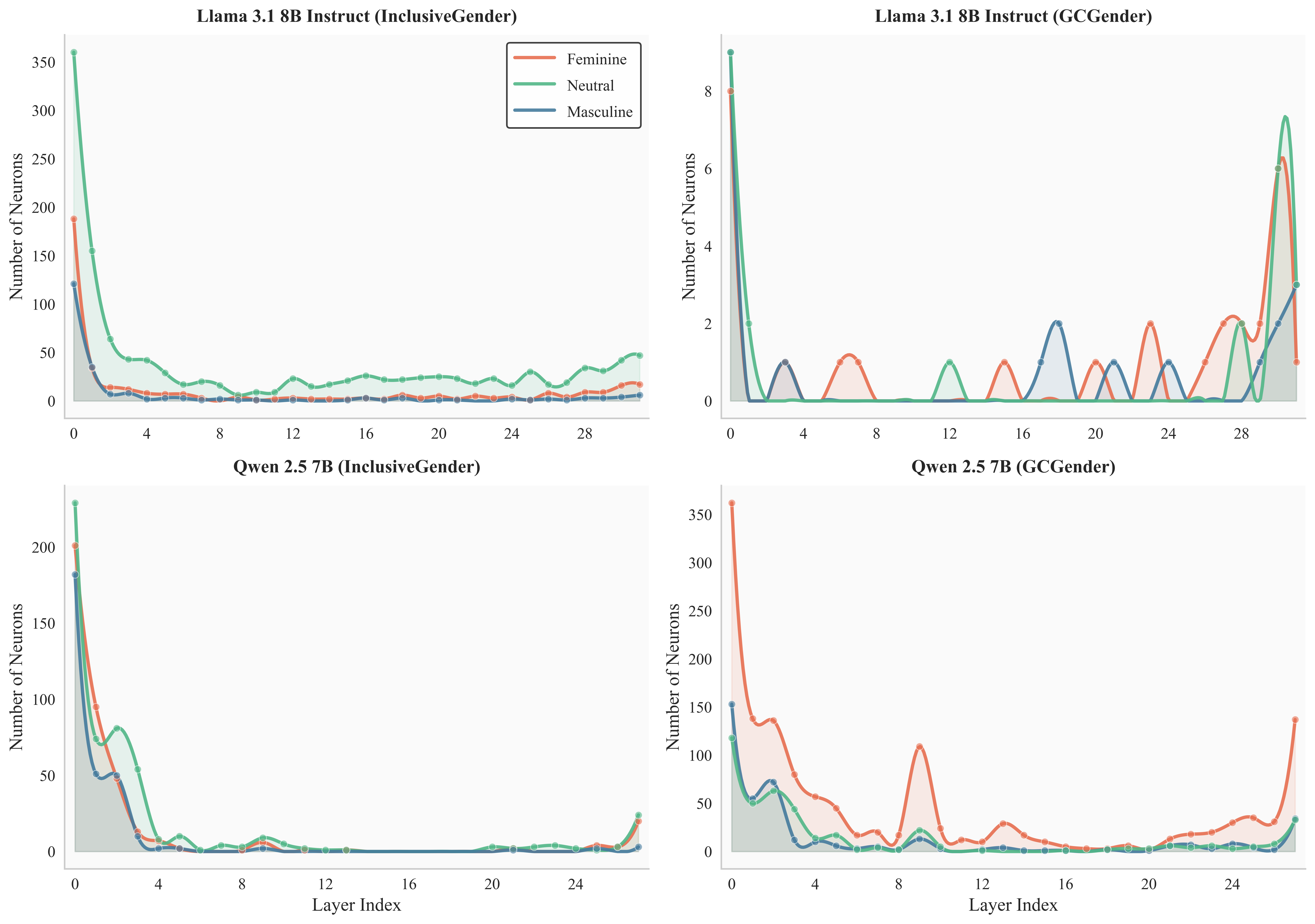}
  \caption{Statistics of the number of gender-specific neurons using our method in each layer in \texttt{Llama 3.1} and \texttt{Qwen 2.5} on \InclusiveGender and \cggender datasets.}
  \label{fig:neuron-distribution}
  \vspace{-20pt}
\end{wrapfigure}
Figure~\ref{fig:neuron-distribution} shows that the gender-related neurons found by our method are not spread evenly across layers. Instead, they concentrate in a few parts of the model. For \InclusiveGender, both \smash{\texttt{Llama 3.1}} and \smash{\texttt{Qwen 2.5}} have a very strong peak in the first 1 to 3 layers, then the counts drop quickly and stay low in the middle layers. This pattern suggests that many gender cues used in rewriting (e.g., pronouns or gendered job titles) are captured in early model layers. We also observe a small but consistent rise in later layers, especially for the neutral set in \texttt{Llama 3.1}, which may reflect that later layers help shape the final wording style.

For \cggender, the two models behave differently. \texttt{Llama 3.1} shows very few selected neurons overall, with most of them appearing in the first layer and the last few layers, indicating that gender control is more localized under this dataset. In contrast, \texttt{Qwen 2.5} shows a much larger set for the feminine direction, again concentrated early, with additional spikes in middle or late layers. This asymmetry indicates that the model may rely more on certain neuron groups when producing feminine forms, while masculine and neutral changes require fewer dedicated neurons.

Overall, our gender neuron identification approach shows more gender-related neurons are in early layers, while competing methods tend to identify more neurons in later layers (see Appendix~\ref{appx:neuron-distribution} for more details), which produce worse performance in neuron masking for causal validation. These distributions support our main finding: our method isolates compact, layer-specific gender neurons, and it reveals clear differences between models and datasets in where gender-related behavior is encoded.

\begin{table*}[th]
\centering
\footnotesize
\resizebox{0.91\textwidth}{!}{%
\begin{tabular}{ll|ccc|ccc|ccc}
\toprule
& & \multicolumn{3}{c|}{\textbf{sNeuron-TST}} & \multicolumn{3}{c|}{\textbf{LAPE}} & \multicolumn{3}{c}{\textbf{Ours}} \\
\cmidrule(lr){3-5}\cmidrule(lr){6-8}\cmidrule(lr){9-11}
\textbf{Dataset} & \textbf{Target} & $\Delta$ M & $\Delta$ F & $\Delta$ N & $\Delta$ M & $\Delta$ F & $\Delta$ N & $\Delta$ M & $\Delta$ F & $\Delta$ N \\
\midrule
\multicolumn{11}{c}{\textbf{Qwen 2.5}} \\
\midrule
%% InclusiveGender - Feminine: Target=F. Ours: F+0.17, M-0.04, N-0.09 -> best (target up, both non-target down)
\multirow{3}{*}{\InclusiveGender}
 & Feminine  & -0.09 & -3.80 & +2.14 & +0.07 & -4.00 & +0.23 & \cellcolor[HTML]{D5F5E3}\textbf{-0.04} & \cellcolor[HTML]{D5F5E3}\textbf{+0.17} & \cellcolor[HTML]{D5F5E3}\textbf{-0.09} \\
%% InclusiveGender - Masculine: Target=M. Ours: M+0.02, F-0.03, N-0.03 -> best (target up, both non-target down)
 & Masculine & +0.17 & +0.07 & +2.23 & -2.03 & -0.11 & +0.26 & \cellcolor[HTML]{D5F5E3}\textbf{+0.02} & \cellcolor[HTML]{D5F5E3}\textbf{-0.03} & \cellcolor[HTML]{D5F5E3}\textbf{-0.03} \\
%% InclusiveGender - Neutral: Target=N. Ours: N+0.22, M-0.03, F-0.01 -> best (target up, both non-target down)
 & Neutral   & +1.11 & +0.30 & +3.40 & -0.16 & +0.06 & +5.81 & \cellcolor[HTML]{D5F5E3}\textbf{-0.03} & \cellcolor[HTML]{D5F5E3}\textbf{-0.01} & \cellcolor[HTML]{D5F5E3}\textbf{+0.22} \\
\midrule
%% CGGender - Feminine: Target=F. Ours: F+0.07, M-0.01, N-0.05 -> best
\multirow{3}{*}{\cggender}
 & Feminine  & +0.02 & -1.53 & -0.34 & -0.06 & -1.83 & +0.13 & \cellcolor[HTML]{D5F5E3}\textbf{-0.01} & \cellcolor[HTML]{D5F5E3}\textbf{+0.07} & \cellcolor[HTML]{D5F5E3}\textbf{-0.05} \\
%% CGGender - Masculine: Target=M. Ours: M+0.06, F-0.11, N-0.02 -> best
 & Masculine & -0.83 & -0.54 & +3.39 & -0.57 & -0.38 & -0.16 & \cellcolor[HTML]{D5F5E3}\textbf{+0.06} & \cellcolor[HTML]{D5F5E3}\textbf{-0.11} & \cellcolor[HTML]{D5F5E3}\textbf{-0.02} \\
%% CGGender - Neutral: Target=N. Ours: N-0.01, M-0.02, F-0.04 -> target drops slightly; LAPE: N-1.05 worse; sNeuron: N+1.61 but M+0.24 leakage. None perfect. Ours has least leakage overall.
 & Neutral   & +0.24 & -0.04 & +1.61 & -0.00 & +0.02 & -1.05 & \cellcolor[HTML]{FFF9C4}\textbf{-0.02} & \cellcolor[HTML]{FFF9C4}\textbf{-0.04} & \cellcolor[HTML]{FFF9C4}\textbf{-0.01} \\
\midrule
\multicolumn{11}{c}{\textbf{{Llama 3.1}}} \\
\midrule
%% InclusiveGender - Feminine: Target=F. Ours: F+0.89, M-0.33, N-0.04 -> best
\multirow{3}{*}{\InclusiveGender}
 & Feminine  & +3.54 & -4.37 & -0.02 & -0.36 & -0.19 & +0.04 & \cellcolor[HTML]{D5F5E3}\textbf{-0.33} & \cellcolor[HTML]{D5F5E3}\textbf{+0.89} & \cellcolor[HTML]{D5F5E3}\textbf{-0.04} \\
%% InclusiveGender - Masculine: Target=M. Ours: M+0.33, F-0.65, N-0.10 -> best
 & Masculine & -2.37 & +1.87 & +0.08 & -1.65 & +0.59 & +0.42 & \cellcolor[HTML]{D5F5E3}\textbf{+0.33} & \cellcolor[HTML]{D5F5E3}\textbf{-0.65} & \cellcolor[HTML]{D5F5E3}\textbf{-0.10} \\
%% InclusiveGender - Neutral: Target=N. Ours: N-0.31, M-0.02, F+0.00 -> target drops. LAPE: N-0.63 worse. sNeuron: N-10.42 much worse. Ours least bad.
 & Neutral   & +0.17 & +0.25 & -10.42 & +0.02 & +0.01 & -0.63 & \cellcolor[HTML]{FFF9C4}\textbf{-0.02} & \cellcolor[HTML]{FFF9C4}\textbf{+0.00} & \cellcolor[HTML]{FFF9C4}\textbf{-0.31} \\
\midrule
%% CGGender - Feminine: Target=F. Ours: F+0.10, M-0.02, N-0.01 -> best
\multirow{3}{*}{\cggender}
 & Feminine  & +2.45 & -4.14 & +0.45 & -0.02 & -0.33 & +0.15 & \cellcolor[HTML]{D5F5E3}\textbf{-0.02} & \cellcolor[HTML]{D5F5E3}\textbf{+0.10} & \cellcolor[HTML]{D5F5E3}\textbf{-0.01} \\
%% CGGender - Masculine: Target=M. Ours: M+0.76, F-0.38, N-0.02 -> best
 & Masculine & -0.92 & -0.27 & -0.34 & -0.62 & -0.03 & +0.07 & \cellcolor[HTML]{D5F5E3}\textbf{+0.76} & \cellcolor[HTML]{D5F5E3}\textbf{-0.38} & \cellcolor[HTML]{D5F5E3}\textbf{-0.02} \\
%% CGGender - Neutral: Target=N. Ours: all 0.00 -> no change at all. sNeuron: N-1.16 drops. LAPE: N-0.46 drops. Ours at least doesn't hurt.
 & Neutral   & +0.04 & -0.04 & -1.16 & +0.02 & -0.03 & -0.46 & \cellcolor[HTML]{FFF9C4}\textbf{+0.00} & \cellcolor[HTML]{FFF9C4}\textbf{+0.00} & \cellcolor[HTML]{FFF9C4}\textbf{+0.00} \\
\bottomrule
\end{tabular}%
}
\caption{Gender neuron masking results ($\Delta$ Ratio, percentage points) for \texttt{Qwen 2.5} and \texttt{Llama 3.1} on \InclusiveGender and \cggender. Each value reports the change in gendered term mention ratio relative to the Baseline under the same target (positive = increase, negative = decrease). For our method, \smash{\setlength{\fboxsep}{1.2pt}\colorbox[HTML]{D5F5E3}{green}} indicates rows where our method achieves the best target-consistent control (target ratio preserved/increased, non-target ratios decreased), while \smash{\setlength{\fboxsep}{1.2pt}\colorbox[HTML]{FFF9C4}{yellow}} marks rows where no method fully succeeds but ours shows the least leakage.}
\label{tab:merged_comparison}
\end{table*}

\subsection{Gender Neuron Masking for Causal Validation} \label{sec:main-results}

We evaluate how well our method identifies \emph{gender-related neurons} by testing whether we can control gendered wording in generation with minimal side effects.
Following the prompting setup of sNeuron-TST \citep{lai2024style}, we prompt the LM to rewrite each test instance into a sentence in a specified target gender.
As a baseline, we transform the input sentences without any neuron intervention.
Then, for each target gender $g$, we apply a \emph{keep-only} intervention: we keep the neurons identified for $g$ active and mask (deactivate) the neurons identified for the other two genders. We measure the change in gendered-term mention ratios relative to the baseline in the same dataset. We hypothesize that the target gender ratio should stay similar or increase, while the other two ratios should drop given their corresponding neuron sets are deactivated.

Table~\ref{tab:merged_comparison} reports the change in gendered-term mention ratios ($\Delta$ Ratio) relative to the baseline for both \texttt{Qwen 2.5} and \texttt{Llama 3.1} across all matched settings where we \emph{keep only} one gender's neurons and set the generation target to the same gender (full per-method counts and ratios are provided in Appendix~\ref{appx:full-tables}).
Across both datasets and both models, our method provides the most stable and selective gender control: the target gender ratio is preserved or slightly increased, while the other two gender ratios decrease or remain near zero. For example, on \InclusiveGender with \texttt{Qwen 2.5}, keeping only feminine neurons increases the feminine ratio (+0.17) while reducing masculine and neutral ratios (-0.04, -0.09). Keeping only masculine neurons slightly increases the masculine ratio (+0.02) with small decreases in the other two (-0.03, -0.03). Keeping only neutral neurons increases the neutral ratio (+0.22) while decreasing masculine and feminine (-0.03, -0.01). 

On \cggender, we observe the same trend for feminine and masculine targets. For the neutral setting, no method fully succeeds; however, ours shows the least leakage (-0.02, -0.04, -0.01). LAPE decreases the target neutral ratio (-1.05) while increasing feminine (+0.02), and although sNeuron-TST increases the target neutral ratio (+1.61), it also increases masculine (+0.24). Results on \texttt{Llama 3.1} follow the same trends, including the neutral setting where no method fully succeeds. Compared to \texttt{Qwen 2.5}, \texttt{Llama 3.1} shows stronger improvements for some targets (e.g., +0.76 masculine on \cggender), while \texttt{Qwen 2.5} shows smaller but steadier gains across all targets. This suggests that models may encode gender control differently, but in both cases our method provides more precise masking than the competing methods.

Competing methods exhibit larger non-target drift. sNeuron-TST frequently inflates unrelated ratios (e.g., +2.23 neutral when targeting masculine on \InclusiveGender for \texttt{Qwen 2.5}), while LAPE shows similar leakage (e.g., +0.26 neutral when targeting masculine on the same dataset), indicating less precise neuron identification.
Overall, results on both models confirm that our method enables more selective gender control with minimal leakage. We also note that training data size affects identification quality: on \cggender (fewer examples), the neutral setting shows inconsistent changes across all methods, suggesting that smaller training sets yield noisier neuron sets. This effect is also visible in the number of identified neurons, especially for \texttt{Llama 3.1} (Figure~\ref{fig:neuron-distribution}). The ablation study of our neuron identification threshold is reported in Appendix~\ref{appx:ablation}.

\subsection{Gender Transformation Quality Evaluation} \label{quality-eval}
Following the evaluation setup described in Section~\ref{sec:eval}, we conduct a human annotation study comparing the baseline (no masking) and our neuron intervention method on \textit{Overall Idea Preservation} and \textit{Target Gender Realization}. We randomly sample 200 instances across target genders for both conditions. Table~\ref{tab:human-eval} reports the percentage of positive labels and pairwise agreement between the two annotators. 

For \textit{Overall Idea Preservation}, the mean positive rate decreases slightly from the baseline to our method, with a drop of 2.75 percentage points. This difference is primarily due to annotator disagreement: Annotator~1 reports near-perfect preservation for our method (100\% vs.\ 98\%), whereas Annotator~2 assigns a lower score (85.9\% vs.\ 93.4\%). This discrepancy likely reflects subjectivity in judging meaning under gender modifications, where neuron intervention can introduce phrasing changes perceived as subtle semantic shifts despite preserving the core content. For \textit{Target Gender Realization}, our method outperforms the baseline, with both annotators independently assigning higher scores (improvements of over seven percentage points). Overall, these results indicate that our neuron intervention method achieves better gender realization with only a minor and annotator-dependent trade-off in meaning preservation. Qualitative examples in Appendix~\ref{appx:case-study} show that neuron-level interventions produce more consistent gender transformations than the baseline.

\begin{table}[t]
\centering
\footnotesize
\resizebox{0.85\textwidth}{!}{%
\setlength{\tabcolsep}{4pt}
\begin{tabular}{lcccccc}
\toprule
 & \multicolumn{3}{c}{\textbf{Baseline}} & \multicolumn{3}{c}{\textbf{Our Method}} \\
\cmidrule(lr){2-4} \cmidrule(lr){5-7}
Criterion & Ann. 1 & Ann. 2 & Agr.($\uparrow$) & Ann. 1 & Ann. 2 & Agr.($\uparrow$) \\
\midrule
Overall Idea Preservation ($\uparrow$) & 98.0 & 93.4 & 95.5 & 100.0 & 85.9 & 85.9 \\
Target Gender Realization ($\uparrow$) & 85.4 & 86.9 & 98.5 & 99.0 & 94.4 & 94.4 \\
\bottomrule
\end{tabular}
}
\caption{Human evaluation results comparing Baseline and Our Method ($N=200$). Annotator 1 and Annotator 2 report the percentage of “Yes” labels for each criterion; Agr. indicates pairwise agreement. $\uparrow$ indicates higher is better.}
\label{tab:human-eval}
\end{table}

\section{Conclusion}
In this paper, we introduce a novel method for identifying and steering gender-specific neurons within language models. Unlike previous studies that focused on a binary view of gender, our work incorporates masculine, feminine, and gender-neutral categories. Experimental results on two sentence-level gendered datasets show that gender representations are encoded in a very small fraction of the model layers, typically less than 0.5\%. These neurons are not spread evenly but are concentrated primarily in the early layers of the model, with smaller contributions from later layers. By using a combined exclusivity score, we identify gender neurons that are specifically tied to one gender rather than general language patterns. Through causal intervention experiments, we show that masking these gender-specific neurons effectively steers the model's output. Compared to existing methods, our approach is more precise and maintains better text generation quality, providing a clearer understanding of how LMs process gendered information.

\section*{Limitations}
(1) We evaluate on two model families (\texttt{Llama 3.1} and \texttt{Qwen 2.5}) at the 7--8B scale. While two distinct architectures support generalizability, we do not test larger variants (e.g., 70B) due to memory constraints.
(2) Our approach may select non-gender-exclusive neurons, especially for the neutral category, though the identified neurons consistently outperform competing methods.
(3) Gender neurons are identified from a fixed set of training sentences and evaluated using a keyword lexicon, which may miss subtler forms of gender bias. We address this with human evaluation and validate across two datasets.
(4) Smaller training sets can introduce noisier neurons, reducing matching precision. Future work should explore data scaling and stronger exclusivity filtering.

\section*{Ethical Considerations}

This work studies gender-related representations in LMs with the goal of improving understanding and controllability of gendered and gender-neutral language generation. We focus on analyzing internal model components rather than deploying new generative systems. The datasets used in this study are derived from publicly available resources or generated by LMs under controlled prompts, and they do not contain personal data or references to real individuals. LM-generated datasets are validated through human annotation to ensure quality.

\bibliography{colm2026_conference}

@inproceedings{kotek2023gender,
    author = {Kotek, Hadas and Dockum, Rikker and Sun, David},
    title = {Gender bias and stereotypes in Large Language Models},
    year = {2023},
    isbn = {9798400701139},
    publisher = {Association for Computing Machinery},
    address = {New York, NY, USA},
    url = {https://doi.org/10.1145/3582269.3615599},
    doi = {10.1145/3582269.3615599},
    booktitle = {Proceedings of The ACM Collective Intelligence Conference},
    pages = {12–24},
    numpages = {13},
    location = {Delft, Netherlands},
    series = {CI '23}
}

@misc{dong2024disclosure,
      title={Disclosure and Mitigation of Gender Bias in LLMs}, 
      author={Xiangjue Dong and Yibo Wang and Philip S. Yu and James Caverlee},
      year={2024},
      eprint={2402.11190},
      archivePrefix={arXiv},
      primaryClass={cs.CL},
      url={https://arxiv.org/abs/2402.11190}, 
}

@inproceedings{dong2023probing,
title={Probing Explicit and Implicit Gender Bias through {LLM} Conditional Text Generation},
author={Xiangjue Dong and Yibo Wang and Philip Yu and James Caverlee},
booktitle={Socially Responsible Language Modelling Research},
year={2023},
url={https://openreview.net/forum?id=ZDeEYmKYrR}
}

@inproceedings{an2025mutual,
    title = "On the Mutual Influence of Gender and Occupation in {LLM} Representations",
    author = "An, Haozhe  and
      Baumler, Connor  and
      Sancheti, Abhilasha  and
      Rudinger, Rachel",
    booktitle = "Proceedings of the 63rd Annual Meeting of the Association for Computational Linguistics (Volume 1: Long Papers)",
    month = jul,
    year = "2025",
    address = "Vienna, Austria",
    publisher = "Association for Computational Linguistics",
    url = "https://aclanthology.org/2025.acl-long.83/",
    doi = "10.18653/v1/2025.acl-long.83",
    pages = "1663--1680",
    ISBN = "979-8-89176-251-0",
}

@inproceedings{thakur2023language,
    title = "Language Models Get a Gender Makeover: Mitigating Gender Bias with Few-Shot Data Interventions",
    author = "Thakur, Himanshu  and
      Jain, Atishay  and
      Vaddamanu, Praneetha  and
      Liang, Paul Pu  and
      Morency, Louis-Philippe",
    editor = "Rogers, Anna  and
      Boyd-Graber, Jordan  and
      Okazaki, Naoaki",
    booktitle = "Proceedings of the 61st Annual Meeting of the Association for Computational Linguistics (Volume 2: Short Papers)",
    month = jul,
    year = "2023",
    address = "Toronto, Canada",
    publisher = "Association for Computational Linguistics",
    url = "https://aclanthology.org/2023.acl-short.30/",
    doi = "10.18653/v1/2023.acl-short.30",
    pages = "340--351",
}

@inproceedings{ma2024debiasing,
    title = "Debiasing Large Language Models with Structured Knowledge",
    author = "Ma, Congda  and
      Zhao, Tianyu  and
      Okumura, Manabu",
    booktitle = "Findings of the Association for Computational Linguistics: ACL 2024",
    month = aug,
    year = "2024",
    address = "Bangkok, Thailand",
    publisher = "Association for Computational Linguistics",
    url = "https://aclanthology.org/2024.findings-acl.612/",
    doi = "10.18653/v1/2024.findings-acl.612",
    pages = "10274--10287"}

@inproceedings{lai2024style,
    title = "Style-Specific Neurons for Steering {LLM}s in Text Style Transfer",
    author = "Lai, Wen  and
      Hangya, Viktor  and
      Fraser, Alexander",
    editor = "Al-Onaizan, Yaser  and
      Bansal, Mohit  and
      Chen, Yun-Nung",
    booktitle = "Proceedings of the 2024 Conference on Empirical Methods in Natural Language Processing",
    month = nov,
    year = "2024",
    address = "Miami, Florida, USA",
    publisher = "Association for Computational Linguistics",
    url = "https://aclanthology.org/2024.emnlp-main.745/",
    doi = "10.18653/v1/2024.emnlp-main.745",
    pages = "13427--13443"}

@inproceedings{oba2024contextual,
    title = "In-Contextual Gender Bias Suppression for Large Language Models",
    author = "Oba, Daisuke  and
      Kaneko, Masahiro  and
      Bollegala, Danushka",
    booktitle = "Findings of the Association for Computational Linguistics: EACL 2024",
    month = mar,
    year = "2024",
    address = "St. Julian{'}s, Malta",
    publisher = "Association for Computational Linguistics",
    url = "https://aclanthology.org/2024.findings-eacl.121/",
    pages = "1722--1742",
}

@inproceedings{tang2024language,
    title = "Language-Specific Neurons: The Key to Multilingual Capabilities in Large Language Models",
    author = "Tang, Tianyi  and
      Luo, Wenyang  and
      Huang, Haoyang  and
      Zhang, Dongdong  and
      Wang, Xiaolei  and
      Zhao, Xin  and
      Wei, Furu  and
      Wen, Ji-Rong",
    editor = "Ku, Lun-Wei  and
      Martins, Andre  and
      Srikumar, Vivek",
    booktitle = "Proceedings of the 62nd Annual Meeting of the Association for Computational Linguistics (Volume 1: Long Papers)",
    month = aug,
    year = "2024",
    address = "Bangkok, Thailand",
    publisher = "Association for Computational Linguistics",
    url = "https://aclanthology.org/2024.acl-long.309/",
    doi = "10.18653/v1/2024.acl-long.309",
    pages = "5701--5715",
}

@inproceedings{kargaran-etal-2025-programming,
    title = "How Programming Concepts and Neurons Are Shared in Code Language Models",
    author = "Kargaran, Amir Hossein  and
      Liu, Yihong  and
      Yvon, Fran{\c{c}}ois  and
      Schuetze, Hinrich",
    booktitle = "Findings of the Association for Computational Linguistics: ACL 2025",
    month = jul,
    year = "2025",
    address = "Vienna, Austria",
    publisher = "Association for Computational Linguistics",
    url = "https://aclanthology.org/2025.findings-acl.1379/",
    doi = "10.18653/v1/2025.findings-acl.1379",
    pages = "26905--26917",
    ISBN = "979-8-89176-256-5"}

@inproceedings{liu2025relation,
    title = "On Relation-Specific Neurons in Large Language Models",
    author = "Liu, Yihong  and
      Chen, Runsheng  and
      Hirlimann, Lea  and
      Hakimi, Ahmad Dawar  and
      Wang, Mingyang  and
      Kargaran, Amir Hossein  and
      Rothe, Sascha  and
      Yvon, Fran{\c{c}}ois  and
      Schuetze, Hinrich",
    booktitle = "Proceedings of the 2025 Conference on Empirical Methods in Natural Language Processing",
    month = nov,
    year = "2025",
    address = "Suzhou, China",
    publisher = "Association for Computational Linguistics",
    url = "https://aclanthology.org/2025.emnlp-main.52/",
    doi = "10.18653/v1/2025.emnlp-main.52",
    pages = "992--1022",
    ISBN = "979-8-89176-332-6"}

@inproceedings{yang2024mitigating,
    title = "Mitigating Biases for Instruction-following Language Models via Bias Neurons Elimination",
    author = "Yang, Nakyeong  and
      Kang, Taegwan  and
      Choi, Stanley Jungkyu  and
      Lee, Honglak  and
      Jung, Kyomin",
    booktitle = "Proceedings of the 62nd Annual Meeting of the Association for Computational Linguistics (Volume 1: Long Papers)",
    month = aug,
    year = "2024",
    address = "Bangkok, Thailand",
    publisher = "Association for Computational Linguistics",
    url = "https://aclanthology.org/2024.acl-long.490/",
    doi = "10.18653/v1/2024.acl-long.490",
    pages = "9061--9073",
}

@inproceedings{
limisiewiczdebiasing,
title={Debiasing Algorithm through Model Adaptation},
author={Tomasz Limisiewicz and David Mare{\v{c}}ek and Tom{\'a}{\v{s}} Musil},
booktitle={The Twelfth International Conference on Learning Representations},
year={2024},
url={https://openreview.net/forum?id=XIZEFyVGC9}
}

@misc{liu2024devil,
      title={The Devil is in the Neurons: Interpreting and Mitigating Social Biases in Pre-trained Language Models}, 
      author={Yan Liu and Yu Liu and Xiaokang Chen and Pin-Yu Chen and Daoguang Zan and Min-Yen Kan and Tsung-Yi Ho},
      year={2024},
      eprint={2406.10130},
      archivePrefix={arXiv},
      primaryClass={cs.CL},
      url={https://arxiv.org/abs/2406.10130}, 
}

@inproceedings{xu2025biasedit,
    title = "{B}ias{E}dit: Debiasing Stereotyped Language Models via Model Editing",
    author = "Xu, Xin  and
      Xu, Wei  and
      Zhang, Ningyu  and
      McAuley, Julian",
    booktitle = "Proceedings of the 5th Workshop on Trustworthy NLP (TrustNLP 2025)",
    month = may,
    year = "2025",
    address = "Albuquerque, New Mexico",
    publisher = "Association for Computational Linguistics",
    url = "https://aclanthology.org/2025.trustnlp-main.13/",
    doi = "10.18653/v1/2025.trustnlp-main.13",
    pages = "166--184",
    ISBN = "979-8-89176-233-6",
}

@inproceedings{qian2024dean,
    title = "The Tug of War Within: Mitigating the Fairness-Privacy Conflicts in Large Language Models",
    author = "Qian, Chen  and
      Liu, Dongrui  and
      Zhang, Jie  and
      Liu, Yong  and
      Shao, Jing",
    booktitle = "Proceedings of the 63rd Annual Meeting of the Association for Computational Linguistics (Volume 1: Long Papers)",
    month = jul,
    year = "2025",
    address = "Vienna, Austria",
    publisher = "Association for Computational Linguistics",
    url = "https://aclanthology.org/2025.acl-long.590/",
    doi = "10.18653/v1/2025.acl-long.590",
    pages = "12066--12095",
    ISBN = "979-8-89176-251-0",
}

@inproceedings{cyberey2025sensing,
    title = "Unsupervised Concept Vector Extraction for Bias Control in {LLM}s",
    author = "Cyberey, Hannah  and
      Ji, Yangfeng  and
      Evans, David",
    editor = "Christodoulopoulos, Christos  and
      Chakraborty, Tanmoy  and
      Rose, Carolyn  and
      Peng, Violet",
    booktitle = "Proceedings of the 2025 Conference on Empirical Methods in Natural Language Processing",
    month = nov,
    year = "2025",
    address = "Suzhou, China",
    publisher = "Association for Computational Linguistics",
    url = "https://aclanthology.org/2025.emnlp-main.1439/",
    doi = "10.18653/v1/2025.emnlp-main.1439",
    pages = "28321--28343",
    ISBN = "979-8-89176-332-6",
}

@inproceedings{vaswani2017attention,
 author = {Vaswani, Ashish and Shazeer, Noam and Parmar, Niki and Uszkoreit, Jakob and Jones, Llion and Gomez, Aidan N and Kaiser, \L ukasz and Polosukhin, Illia},
 booktitle = {Advances in Neural Information Processing Systems},
 editor = {I. Guyon and U. Von Luxburg and S. Bengio and H. Wallach and R. Fergus and S. Vishwanathan and R. Garnett},
 pages = {},
 publisher = {Curran Associates, Inc.},
 title = {Attention is All you Need},
 url = {https://proceedings.neurips.cc/paper_files/paper/2017/file/3f5ee243547dee91fbd053c1c4a845aa-Paper.pdf},
 volume = {30},
 year = {2017}
}

@INPROCEEDINGS{soundararajan2023using,
  author={Soundararajan, Shweta and Jeyaraj, Manuela Nayantara and Delany, Sarah Jane},
  booktitle={2023 31st Irish Conference on Artificial Intelligence and Cognitive Science (AICS)}, 
  title={Using ChatGPT to Generate Gendered Language}, 
  year={2023},
  volume={},
  number={},
  pages={1-8},
  doi={10.1109/AICS60730.2023.10470830},
  url={https://ieeexplore.ieee.org/document/10470830}
  }

@article{gaucher2011evidence,
  title={Evidence that gendered wording in job advertisements exists and sustains gender inequality.},
  author={Gaucher, Danielle and Friesen, Justin and Kay, Aaron C},
  journal={Journal of personality and social psychology},
  volume={101},
  number={1},
  pages={109},
  year={2011},
  url={https://psycnet.apa.org/record/2011-04642-001},
  publisher={American Psychological Association}
}

@inproceedings{cryan2020detecting,
    author = {Cryan, Jenna and Tang, Shiliang and Zhang, Xinyi and Metzger, Miriam and Zheng, Haitao and Zhao, Ben Y.},
    title = {Detecting Gender Stereotypes: Lexicon vs. Supervised Learning Methods},
    year = {2020},
    isbn = {9781450367080},
    publisher = {Association for Computing Machinery},
    address = {New York, NY, USA},
    url = {https://doi.org/10.1145/3313831.3376488},
    doi = {10.1145/3313831.3376488},
    booktitle = {Proceedings of the 2020 CHI Conference on Human Factors in Computing Systems},
    pages = {1–11},
    numpages = {11},
    location = {Honolulu, HI, USA},
    series = {CHI '20}
}

@misc{henderson_inclusive_language,
  author = {Henderson, Joel Parker},
  title = {Inclusive Language},
  year = {2023},
  publisher = {GitHub},
  howpublished = {\url{https://github.com/joelparkerhenderson/inclusive-language}},
}

@article{elfwing2018sigmoid,
    title = {Sigmoid-weighted linear units for neural network function approximation in reinforcement learning},
    journal = {Neural Networks},
    volume = {107},
    pages = {3-11},
    year = {2018},
    note = {Special issue on deep reinforcement learning},
    issn = {0893-6080},
    doi = {https://doi.org/10.1016/j.neunet.2017.12.012},
    url = {https://www.sciencedirect.com/science/article/pii/S0893608017302976},
    author = {Stefan Elfwing and Eiji Uchibe and Kenji Doya},
}

@inproceedings{you2024beyond,
    title = "Beyond Binary Gender Labels: Revealing Gender Bias in {LLM}s through Gender-Neutral Name Predictions",
    author = "You, Zhiwen  and
      Lee, HaeJin  and
      Mishra, Shubhanshu  and
      Jeoung, Sullam  and
      Mishra, Apratim  and
      Kim, Jinseok  and
      Diesner, Jana",
    booktitle = "Proceedings of the 5th Workshop on Gender Bias in Natural Language Processing (GeBNLP)",
    month = aug,
    year = "2024",
    address = "Bangkok, Thailand",
    publisher = "Association for Computational Linguistics",
    url = "https://aclanthology.org/2024.gebnlp-1.16/",
    doi = "10.18653/v1/2024.gebnlp-1.16",
    pages = "255--268",
}

@inproceedings{nikeghbal-etal-2025-cobia,
    title = "{C}o{B}ia: Constructed Conversations Can Trigger Otherwise Concealed Societal Biases in {LLM}s",
    author = "Nikeghbal, Nafiseh  and
      Kargaran, Amir Hossein  and
      Diesner, Jana",
    booktitle = "Proceedings of the 2025 Conference on Empirical Methods in Natural Language Processing",
    month = nov,
    year = "2025",
    address = "Suzhou, China",
    publisher = "Association for Computational Linguistics",
    url = "https://aclanthology.org/2025.emnlp-main.84/",
    doi = "10.18653/v1/2025.emnlp-main.84",
    pages = "1618--1639",
    ISBN = "979-8-89176-332-6"}

@inproceedings{you2024sciprompt,
  title={SciPrompt: Knowledge-augmented prompting for fine-grained categorization of scientific topics},
  author={You, Zhiwen and Han, Kanyao and Zhu, Haotian and Lud{\"a}scher, Bertram and Diesner, Jana},
  booktitle={Proceedings of the 2024 Conference on Empirical Methods in Natural Language Processing},
  pages={6087--6104},
  year={2024},
  url={https://aclanthology.org/2024.emnlp-main.350/}
}

@misc{lee2025revisiting,
      title={Revisiting gender bias research in bibliometrics: Standardizing methodological variability using Scholarly Data Analysis (SoDA) Cards}, 
      author={HaeJin Lee and Shubhanshu Mishra and Apratim Mishra and Zhiwen You and Jinseok Kim and Jana Diesner},
      year={2025},
      eprint={2501.18129},
      archivePrefix={arXiv},
      primaryClass={cs.DL},
      url={https://arxiv.org/abs/2501.18129}, 
}

@misc{gpt4o,
  author       = {{OpenAI}},
  title        = {GPT-4o},
  year         = {2024},
  howpublished = {\url{https://platform.openai.com/docs/models/gpt-4o}}
}

@inproceedings{kojima-etal-2024-multilingual,
    title = "On the Multilingual Ability of Decoder-based Pre-trained Language Models: Finding and Controlling Language-Specific Neurons",
    author = "Kojima, Takeshi  and
      Okimura, Itsuki  and
      Iwasawa, Yusuke  and
      Yanaka, Hitomi  and
      Matsuo, Yutaka",
    booktitle = "Proceedings of the 2024 Conference of the North American Chapter of the Association for Computational Linguistics: Human Language Technologies (Volume 1: Long Papers)",
    month = jun,
    year = "2024",
    address = "Mexico City, Mexico",
    publisher = "Association for Computational Linguistics",
    url = "https://aclanthology.org/2024.naacl-long.384/",
    doi = "10.18653/v1/2024.naacl-long.384",
    pages = "6919--6971",
}

@misc{wang2024sharing,
      title={Sharing Matters: Analysing Neurons Across Languages and Tasks in LLMs}, 
      author={Weixuan Wang and Barry Haddow and Minghao Wu and Wei Peng and Alexandra Birch},
      year={2025},
      eprint={2406.09265},
      archivePrefix={arXiv},
      primaryClass={cs.CL},
      url={https://arxiv.org/abs/2406.09265}, 
}

@inproceedings{stanczak-etal-2022-neurons,
    title = "Same Neurons, Different Languages: Probing Morphosyntax in Multilingual Pre-trained Models",
    author = "Stanczak, Karolina  and
      Ponti, Edoardo  and
      Torroba Hennigen, Lucas  and
      Cotterell, Ryan  and
      Augenstein, Isabelle",
    booktitle = "Proceedings of the 2022 Conference of the North American Chapter of the Association for Computational Linguistics: Human Language Technologies",
    month = jul,
    year = "2022",
    address = "Seattle, United States",
    publisher = "Association for Computational Linguistics",
    url = "https://aclanthology.org/2022.naacl-main.114/",
    doi = "10.18653/v1/2022.naacl-main.114",
    pages = "1589--1598"}

@inproceedings{zhang-etal-2025-multilingual,
    title = "Multilingual Knowledge Editing with Language-Agnostic Factual Neurons",
    author = "Zhang, Xue  and
      Liang, Yunlong  and
      Meng, Fandong  and
      Zhang, Songming  and
      Chen, Yufeng  and
      Xu, Jinan  and
      Zhou, Jie",
    booktitle = "Proceedings of the 31st International Conference on Computational Linguistics",
    month = jan,
    year = "2025",
    address = "Abu Dhabi, UAE",
    publisher = "Association for Computational Linguistics",
    url = "https://aclanthology.org/2025.coling-main.385/",
    pages = "5775--5788"}

@inproceedings{lutz-etal-2024-local,
    title = "Local Contrastive Editing of Gender Stereotypes",
    author = "Lutz, Marlene  and
      Choenni, Rochelle  and
      Strohmaier, Markus  and
      Lauscher, Anne",
    booktitle = "Proceedings of the 2024 Conference on Empirical Methods in Natural Language Processing",
    month = nov,
    year = "2024",
    address = "Miami, Florida, USA",
    publisher = "Association for Computational Linguistics",
    url = "https://aclanthology.org/2024.emnlp-main.1197/",
    doi = "10.18653/v1/2024.emnlp-main.1197",
    pages = "21474--21493",
}

@misc{llama31_8b_instruct,
  author       = {{Meta AI}},
  title        = {Llama 3.1 8B Instruct},
  year         = {2024},
  howpublished = {\url{https://huggingface.co/meta-llama/Llama-3.1-8B-Instruct}}
}

@misc{qwen2_5_7b,
  author       = {{Qwen}},
  title        = {Qwen2.5-7B},
  year         = {2024},
  howpublished = {\url{https://huggingface.co/Qwen/Qwen2.5-7B}}
}

@misc{qwen2025qwen25technicalreport,
      title={Qwen2.5 Technical Report}, 
      author={Qwen and : and An Yang and Baosong Yang and Beichen Zhang and Binyuan Hui and Bo Zheng and Bowen Yu and Chengyuan Li and Dayiheng Liu and Fei Huang and Haoran Wei and Huan Lin and Jian Yang and Jianhong Tu and Jianwei Zhang and Jianxin Yang and Jiaxi Yang and Jingren Zhou and Junyang Lin and Kai Dang and Keming Lu and Keqin Bao and Kexin Yang and Le Yu and Mei Li and Mingfeng Xue and Pei Zhang and Qin Zhu and Rui Men and Runji Lin and Tianhao Li and Tianyi Tang and Tingyu Xia and Xingzhang Ren and Xuancheng Ren and Yang Fan and Yang Su and Yichang Zhang and Yu Wan and Yuqiong Liu and Zeyu Cui and Zhenru Zhang and Zihan Qiu},
      year={2025},
      eprint={2412.15115},
      archivePrefix={arXiv},
      primaryClass={cs.CL},
      url={https://arxiv.org/abs/2412.15115}, 
}

@misc{grattafiori2024llama3herdmodels,
      title={The Llama 3 Herd of Models}, 
      author={Aaron Grattafiori and Abhimanyu Dubey and Abhinav Jauhri and Abhinav Pandey and Abhishek Kadian and Ahmad Al-Dahle and Aiesha Letman and Akhil Mathur and Alan Schelten and Alex Vaughan and others},
      year={2024},
      eprint={2407.21783},
      archivePrefix={arXiv},
      primaryClass={cs.AI},
      url={https://arxiv.org/abs/2407.21783}, 
}

@misc{openai2024gpt4ocard,
      title={GPT-4o System Card}, 
      author={OpenAI and Aaron Hurst and Adam Lerer and Adam P. Goucher and Adam Perelman and Aditya Ramesh and Aidan Clark and AJ Ostrow and Akila Welihinda and Alan Hayes and Alec Radford and others},
      year={2024},
      eprint={2410.21276},
      archivePrefix={arXiv},
      primaryClass={cs.CL},
      url={https://arxiv.org/abs/2410.21276}, 
}

@misc{llama33_70b_ollama,
  author       = {{Meta AI}},
  title        = {Llama 3.3 70B (Ollama Library)},
  year         = {2024},
  howpublished = {\url{https://ollama.com/library/llama3.3:70b}},
  note         = {Accessed: 2026-03}
}

@inproceedings{piergentili-etal-2023-gender,
    title = "Gender Neutralization for an Inclusive Machine Translation: from Theoretical Foundations to Open Challenges",
    author = "Piergentili, Andrea  and
      Fucci, Dennis  and
      Savoldi, Beatrice  and
      Bentivogli, Luisa  and
      Negri, Matteo",
    editor = "Vanmassenhove, Eva  and
      Savoldi, Beatrice  and
      Bentivogli, Luisa  and
      Daems, Joke  and
      Hackenbuchner, Jani{\c{c}}a",
    booktitle = "Proceedings of the First Workshop on Gender-Inclusive Translation Technologies",
    month = jun,
    year = "2023",
    address = "Tampere, Finland",
    publisher = "European Association for Machine Translation",
    url = "https://aclanthology.org/2023.gitt-1.7/",
    pages = "71--83",
}

@inproceedings{dawkins-etal-2025-gender,
    title = "Gender-Neutral Machine Translation Strategies in Practice",
    author = "Dawkins, Hillary  and
      Nejadgholi, Isar  and
      Lo, Chi-Kiu",
    editor = "Hackenbuchner, Jani{\c{c}}a  and
      Bentivogli, Luisa  and
      Daems, Joke  and
      Manna, Chiara  and
      Savoldi, Beatrice  and
      Vanmassenhove, Eva",
    booktitle = "Proceedings of the 3rd Workshop on Gender-Inclusive Translation Technologies (GITT 2025)",
    month = jun,
    year = "2025",
    address = "Geneva, Switzerland",
    publisher = "European Association for Machine Translation",
    url = "https://aclanthology.org/2025.gitt-1.5/",
    pages = "74--88",
    ISBN = "978-2-9701897-4-9",
}

@inproceedings{savoldi-etal-2025-mind,
    title = "Mind the Inclusivity Gap: Multilingual Gender-Neutral Translation Evaluation with m{G}e{NTE}",
    author = "Savoldi, Beatrice  and
      Attanasio, Giuseppe  and
      Cupin, Eleonora  and
      Gkovedarou, Eleni  and
      Hackenbuchner, Jani{\c{c}}a  and
      Lauscher, Anne  and
      Negri, Matteo  and
      Piergentili, Andrea  and
      Thind, Manjinder  and
      Bentivogli, Luisa",
    editor = "Christodoulopoulos, Christos  and
      Chakraborty, Tanmoy  and
      Rose, Carolyn  and
      Peng, Violet",
    booktitle = "Proceedings of the 2025 Conference on Empirical Methods in Natural Language Processing",
    month = nov,
    year = "2025",
    address = "Suzhou, China",
    publisher = "Association for Computational Linguistics",
    url = "https://aclanthology.org/2025.emnlp-main.692/",
    doi = "10.18653/v1/2025.emnlp-main.692",
    pages = "13698--13720",
    ISBN = "979-8-89176-332-6",
}

@inproceedings{manna-etal-2025-paying,
    title = "Are We Paying Attention to Her? Investigating Gender Disambiguation and Attention in Machine Translation",
    author = "Manna, Chiara  and
      Alishahi, Afra  and
      Blain, Fr{\'e}d{\'e}ric  and
      Vanmassenhove, Eva",
    editor = "Hackenbuchner, Jani{\c{c}}a  and
      Bentivogli, Luisa  and
      Daems, Joke  and
      Manna, Chiara  and
      Savoldi, Beatrice  and
      Vanmassenhove, Eva",
    booktitle = "Proceedings of the 3rd Workshop on Gender-Inclusive Translation Technologies (GITT 2025)",
    month = jun,
    year = "2025",
    address = "Geneva, Switzerland",
    publisher = "European Association for Machine Translation",
    url = "https://aclanthology.org/2025.gitt-1.1/",
    pages = "1--16",
    ISBN = "978-2-9701897-4-9",
}

@misc{hackenbuchner2026triggersmodelcontrastiveexplanations,
      title={What Triggers my Model? Contrastive Explanations Inform Gender Choices by Translation Models}, 
      author={Janiça Hackenbuchner and Arda Tezcan and Joke Daems},
      year={2026},
      eprint={2512.08440},
      archivePrefix={arXiv},
      primaryClass={cs.CL},
      url={https://arxiv.org/abs/2512.08440}, 
}

@inproceedings{attanasio-etal-2023-tale,
    title = "A Tale of Pronouns: Interpretability Informs Gender Bias Mitigation for Fairer Instruction-Tuned Machine Translation",
    author = "Attanasio, Giuseppe  and
      Plaza del Arco, Flor Miriam  and
      Nozza, Debora  and
      Lauscher, Anne",
    editor = "Bouamor, Houda  and
      Pino, Juan  and
      Bali, Kalika",
    booktitle = "Proceedings of the 2023 Conference on Empirical Methods in Natural Language Processing",
    month = dec,
    year = "2023",
    address = "Singapore",
    publisher = "Association for Computational Linguistics",
    url = "https://aclanthology.org/2023.emnlp-main.243/",
    doi = "10.18653/v1/2023.emnlp-main.243",
    pages = "3996--4014",
}
\bibliographystyle{colm2026_conference}
\clearpage
\appendix

\section{Ablation Study} \label{appx:ablation}
To validate the robustness of our neuron selection criteria, we conduct a threshold sensitivity analysis on the \cggender dataset by varying the three key selection thresholds: effect size $d_g$, log-odds difference $\Delta_g$, and minimum positive-activation rate $p_g$. For each threshold, we select across multiple candidate values while holding the remaining two fixed at their defaults, and measure the resulting change in gender-term ratios ($\Delta$ Ratio \%) under the feminine target setting (Table~\ref{tab:threshold_sensitivity}).

As we discuss in Section~\ref{sec:main-results}, the target gender ratio (i.e., feminine) should be preserved or slightly increased, while the other two gender ratios decrease or remain near zero. The results demonstrate that the default configuration $d_g = 0.5$, $\Delta_g = 0.7$, and $p_g = 0.08$, consistently achieves the most favorable trade-off: it increases the target (feminine) gender ratio while minimizing leakage into non-target categories (masculine and neutral). These findings confirm that our selected thresholds represent an optimal operating point for gender neuron identification and intervention.
%%%
\begin{table*}[ht]
\centering
\footnotesize
\resizebox{0.9\textwidth}{!}{%
\begin{tabular*}{\textwidth}{@{\extracolsep\fill}llccc}
\toprule
\textbf{Parameter} & \textbf{Value} &
\multicolumn{3}{c}{\textbf{$\Delta$ Ratio (\%)}} \\
\cmidrule(lr){3-5}
& & \textbf{Masculine} & \textbf{Feminine} & \textbf{Neutral} \\
\midrule
\multirow{5}{*}{Effect Size $d_g$}
 & 0.1          & $+0.03$ & $+0.01$ & $+0.03$ \\
 & 0.3          & $+0.05$ & $+0.05$ & $-0.01$ \\
 & \textbf{0.5} & $\mathbf{-0.01}$ & $\mathbf{+0.07}$ & $\mathbf{-0.05}$ \\
 & 0.7          & $+0.00$ & $-0.01$ & $+0.09$ \\
 & 0.9          & $+0.04$ & $+0.02$ & $+0.14$ \\
\midrule
\multirow{5}{*}{Log-Odds Difference $\Delta_g$}
 & 0.1          & $+0.10$ & $-0.01$ & $+0.02$ \\
 & 0.3          & $+0.04$ & $-0.12$ & $-0.03$ \\
 & 0.5          & $+0.07$ & $-0.07$ & $+0.04$ \\
 & \textbf{0.7} & $\mathbf{-0.01}$ & $\mathbf{+0.07}$ & $\mathbf{-0.05}$ \\
 & 0.9          & $+0.02$ & $+0.08$ & $+0.10$ \\
\midrule
\multirow{6}{*}{Min.\ positive-activate rate $p_g$}
 & 0.01          & $+0.00$ & $-0.08$ & $+0.05$ \\
 & 0.03          & $+0.05$ & $-0.05$ & $+0.01$ \\
 & 0.05          & $+0.06$ & $-0.00$ & $-0.04$ \\
 & \textbf{0.08} & $\mathbf{-0.01}$ & $\mathbf{+0.07}$ & $\mathbf{-0.05}$ \\
 & 0.10          & $+0.02$ & $-0.10$ & $+0.06$ \\
 & 0.12          & $+0.04$ & $-0.22$ & $-0.09$ \\
\bottomrule
\end{tabular*}
}
\caption{Threshold sensitivity analysis on \cggender under the feminine target setting. Bold rows indicate the selected default thresholds. $\Delta$ Ratio (\%) reports the change in gender-term ratios for each gender category.}
\label{tab:threshold_sensitivity}
\end{table*}
%%%

%%
\clearpage

\section{Gender Term Lexicon}
\label{app:gender_terms}

In Table~\ref{tab:gender_lexicon}, Table~\ref{tab:gender_female_lexicon}, and Table~\ref{tab:gender_neutral_lexicon}, we report the gender term lexicon to count gender-specific terms in model outputs. Terms are organized into three categories: masculine, feminine, and gender-neutral. Each category includes pronouns, basic terms, titles, family terms, occupational terms, and other related vocabulary.

\begin{table*}[ht]
\centering
\footnotesize
\resizebox{0.9\textwidth}{!}{%
\begin{tabular}{p{2.5cm}p{10.5cm}}
\toprule
\textbf{Category} & \textbf{Terms} \\
\midrule
\multicolumn{2}{c}{\cellcolor{blue!10}\textbf{Masculine Terms}} \\
\midrule
Pronouns & he, him, his, himself \\
\addlinespace
Basic Terms & man, men, male, males, boy, boys, guy, guys \\
\addlinespace
Titles & sir, gentleman, gentlemen, mr, mister \\
\addlinespace
Family & father, dad, daddy, son, sons, brother, brothers, uncle, uncles, nephew, nephews, husband, husbands, groom, grooms, widower, widowers, godfather, godfathers, grandfather, grandpa, grandson, grandsons \\
\addlinespace
Royalty & king, kings, prince, princes, lord, lords, duke, dukes, baron, barons \\
\addlinespace
Occupational (-man/-men) & businessman, businessmen, chairman, chairmen, congressman, congressmen, fireman, firemen, policeman, policemen, mailman, mailmen, salesman, salesmen, spokesman, spokesmen, craftsman, craftsmen, cameraman, cameramen, anchorman, anchormen, weatherman, weathermen, foreman, foremen, freshman, freshmen, serviceman, servicemen, doorman, doormen, postman, postmen, repairman, repairmen, workman, workmen, clergyman, clergymen, statesman, statesmen, councilman, councilmen, alderman, aldermen, assemblyman, assemblymen, patrolman, patrolmen, handyman, handymen, watchman, watchmen, ombudsman, ombudsmen, bondsman, bondsmen, middleman, middlemen, layman, laymen, horseman, horsemen, fisherman, fishermen, deliveryman, deliverymen, newsman, newsmen, pressman, pressmen \\
\addlinespace
Occupational (-boy) & bellboy, busboy, cowboy, cowboys, paperboy, paperboys, schoolboy, schoolboys, choirboy, choirboys, newsboy, newsboys, batboy, batboys, bagboy, bagboys, playboy, playboys, flyboy, flyboys, cabinboy, cabinboys \\
\addlinespace
Other & actor, actors, bachelor, bachelors, alumnus, alumni, host, hosts, waiter, waiters, steward, stewards, patron, patrons, masseur, masseurs, usher, ushers, headmaster, headmasters, landlord, landlords, mankind, manpower, manmade, brotherhood, fatherhood, fatherland, forefathers \\
\bottomrule
\end{tabular}
}
\caption{Masculine term lexicon used for evaluating gendered language in model outputs. The lexicon contains 139 masculine terms.}
\label{tab:gender_lexicon}
\end{table*}

\begin{table*}[ht]
\centering
\footnotesize
\resizebox{0.9\textwidth}{!}{%
\begin{tabular}{p{2.5cm}p{10.5cm}}
\toprule
\textbf{Category} & \textbf{Terms} \\
\midrule
\multicolumn{2}{c}{\cellcolor{red!10}\textbf{Feminine Terms}} \\
\midrule
Pronouns & she, her, hers, herself \\
\addlinespace
Basic Terms & woman, women, female, females, girl, girls, gal, gals \\
\addlinespace
Titles & ma'am, madam, lady, ladies, miss, ms, mrs \\
\addlinespace
Family & mother, mom, mommy, daughter, daughters, sister, sisters, aunt, aunts, niece, nieces, wife, wives, bride, brides, widow, widows, godmother, godmothers, grandmother, grandma, granddaughter, granddaughters \\
\addlinespace
Royalty & queen, queens, princess, princesses, duchess, duchesses, baroness, baronesses \\
\addlinespace
Occupational (-woman/-women) & businesswoman, businesswomen, chairwoman, chairwomen, congresswoman, congresswomen, firewoman, firewomen, policewoman, policewomen, mailwoman, mailwomen, saleswoman, saleswomen, spokeswoman, spokeswomen, craftswoman, craftswomen, camerawoman, camerawomen, anchorwoman, anchorwomen, weatherwoman, weatherwomen, forewoman, forewomen, servicewoman, servicewomen, clergywoman, clergywomen, stateswoman, stateswomen, councilwoman, councilwomen, alderwoman, alderwomen, assemblywoman, assemblywomen, patrolwoman, patrolwomen, fisherwoman, fisherwomen, newswoman, newswomen \\
\addlinespace
Occupational (-girl) & bellgirl, busgirl, cowgirl, cowgirls, papergirl, papergirls, schoolgirl, schoolgirls, choirgirl, choirgirls \\
\addlinespace
Other (-ess/-ette) & actress, actresses, bachelorette, bachelorettes, alumna, alumnae, hostess, hostesses, waitress, waitresses, stewardess, stewardesses, patroness, patronesses, masseuse, masseuses, usherette, usherettes, poetess, poetesses, sorceress, sorceresses, priestess, priestesses, headmistress, headmistresses, landlady, landladies, maid, maids, bridesmaid, bridesmaids, womankind, sisterhood, motherhood, motherland \\
\bottomrule
\end{tabular}
}
\caption{Feminine term lexicon used for evaluating gendered language in model outputs. The lexicon contains 127 feminine terms.}
\label{tab:gender_female_lexicon}
\end{table*}
\begin{table*}[ht]
\centering
\footnotesize
\resizebox{0.9\textwidth}{!}{%
\begin{tabular}{p{2.5cm}p{10.5cm}}
\toprule
\textbf{Category} & \textbf{Terms} \\
\midrule
\multicolumn{2}{c}{\cellcolor{green!10}\textbf{Gender-Neutral Terms}} \\
\midrule
Pronouns & they, them, their, theirs, themselves \\
\addlinespace
Basic Terms & person, persons, people, individual, individuals, human, humans, adult, adults, child, children, youth, youths \\
\addlinespace
Occupational (-person) & businessperson, chairperson, congressperson, spokesperson, salesperson, craftsperson, cameraperson, anchorperson, foreperson, serviceperson, repairperson, clergyperson, statesperson, councilperson, assemblyperson, layperson, fisherperson, newsperson \\
\addlinespace
Occupational (neutral) & firefighter, firefighters, police officer, mail carrier, letter carrier, postal worker, flight attendant, sales representative, camera operator, news anchor, meteorologist, supervisor, manager, executive, professional, technician, specialist, worker, workers, employee, employees, staff, colleague, colleagues, coworker, coworkers, server, servers, attendant, attendants, assistant, assistants \\
\addlinespace
Family & parent, parents, sibling, siblings, spouse, spouses, partner, partners, offspring, guardian, guardians, caregiver, caregivers, godparent, godparents, grandparent, grandparents, grandchild, grandchildren \\
\addlinespace
Royalty & monarch, monarchs, sovereign, sovereigns, royal, royals, noble, nobles, regent, regents, ruler, rulers, heir, heirs \\
\addlinespace
Other & graduate, graduates, alum, alums, student, students, citizen, citizens, resident, residents, member, members, leader, leaders, director, directors, coordinator, coordinators, humanity, humankind, homemaker, homemakers \\
\bottomrule
\end{tabular}
}
\caption{Gender-neutral term lexicon used for evaluating gendered language in model outputs. The lexicon contains 112 gender-neutral terms.}
\label{tab:gender_neutral_lexicon}
\end{table*}

\clearpage

\section{Dataset Construction Prompts} \label{appx:dataset-prompts}
We use two prompts to construct our datasets (Section~\ref{sec:datasets}). 
The \cggender prompt (Figure \ref{fig:cggender-prompt}) generates gender-neutral versions of existing gendered sentences while preserving their meaning.
The \InclusiveGender prompt (Figure \ref{fig:inclusivegender-prompt}) generates sentences for feminine, masculine, and gender-neutral categories using curated gendered and neutral term lists.
\begin{figure}[ht]
\begin{tcolorbox}[
  enhanced,
  breakable,
  boxrule=1pt,
  boxsep=5pt,
  arc=4pt,
  title=\textit{\cggender Prompt}
]
\footnotesize
\noindent You are given a sentence that contains gendered language, pronouns, names, or stereotypically gendered traits. Your task is to rewrite the sentence to make it gender-neutral and non-binary. This includes changing every pronoun, noun, or name that refers to a male or female person into a gender-neutral equivalent, without changing the meaning of the sentence.

- Replace gendered pronouns and nouns with neutral ones such as:  
    - "he" / "she" → "they"  
    - "him" / "her" → "them"  
    - "his" / "hers" → "their" / "theirs"  
    - "man" / "woman" → "person"  
    - "boy" / "girl" → "child"  
    - "son" / "daughter" → "child"  
    - "mother" / "father" → "parent"  
    - "brother" / "sister" → "sibling"  
    - "groom" / "bride" → "betrothed"  
    - "uncle" / "aunt" → "relative"  
    - "waiter" / "waitress" → "server"  
    - "salesman" / "saleswoman" → "salesperson"  
    - "step-father" / "step-mother" → "step-parent"  
    - "policeman" / "policewoman" → "police officer"  
    - Any specific names (e.g., "Mary", "John") → gender-neutral names (e.g., "Alex", "Taylor")  

- Replace any adjectives or job titles that are stereotypically gender-coded with neutral equivalents. Reduce gender associations and avoid reinforcing stereotypes (e.g., "daring boy" → "adventurous child", "emotional woman" → "sentimental person").  

- Return ONLY the rewritten sentence. Do not provide explanations.

\#\#\# Input Sentence:  
\textcolor{blue}{\{sent\}}

\#\#\# Output (Gender-neutral version):
\end{tcolorbox}
\caption{\cggender Prompt}\label{fig:cggender-prompt}
\end{figure}
\begin{figure*}[ht]
\begin{tcolorbox}[
  enhanced,
  breakable,
  boxrule=1pt,
  boxsep=5pt,
  arc=4pt,
  title=\textit{\InclusiveGender Prompt}
]
\footnotesize
\noindent You are a precise and structured text generator. You are given a list of gendered and gender-neutral terms. Using these terms, generate sentences according to the following instructions.

\noindent \textbf{Input Set:} \{gender\_terms\}

\noindent \textbf{Pronoun Set:} 
Gender-exclusive singular pronouns: he, she, he/she, him, her, him/her, his, hers, his/her, himself, herself, himself/herself.  
Gender-inclusive singular pronouns: they, them, their, theirs, themselves, themself, oneself.  

\noindent For each selected term from the Input Set, include at least one appropriate pronoun from the corresponding Pronoun Set.

\noindent \textbf{Tasks:}  
1. Generate three categories of sentences:  
   - Feminine sentences — use gender-exclusive feminine terms and pronouns.  
   - Masculine sentences — use gender-exclusive masculine terms and pronouns.  
   - Gender-neutral sentences — use gender-inclusive terms and pronouns.  

2. For each category:  
   - Write exactly 40 different sentences.  
   - Number each sentence clearly from 1 to 40.  
   - Each sentence must contain exactly one gender term from the Input Set.  
   - Each sentence must contain at least one pronoun from the same gender group.  
   - Sentences must be grammatically correct and natural.  
   - Do NOT include any specific human names.  
   - Sentences must belong to only one gender category — do not mix indicators.  
   - Do NOT summarize or skip numbers. Produce all 40 sentences fully.  
   - If you reach 40 sentences, stop there.  

3. After each sentence, list all gender indicators (the gender term and the pronoun(s) used) under a single key called \texttt{gender\_indicators}.

\noindent \textbf{Output structure (plain text, no JSON):}

\noindent \textless START\_Feminine\textgreater  
1. [sentence]  
   Gender indicators: [term(s), pronoun(s)]  
...  
40. [sentence]  
   Gender indicators: [term(s), pronoun(s)]  
\textless END\_Feminine\textgreater  

\textless START\_Masculine\textgreater  
1. [sentence]  
   Gender indicators: [term(s), pronoun(s)]  
...  
40. [sentence]  
   Gender indicators: [term(s), pronoun(s)]  
\textless END\_Masculine\textgreater  

\textless START\_Gender-neutral\textgreater  
1. [sentence]  
   Gender indicators: [term(s), pronoun(s)]  
...  
40. [sentence]  
   Gender indicators: [term(s), pronoun(s)]  
\textless END\_Gender-neutral\textgreater  

\noindent Do not include any commentary or summaries. Follow the structure above exactly.  

\noindent \textbf{Example (only 2 shown per category — real output must include 40 each):}  

\noindent Input Set:  
Gender-exclusive: ad-man, ad-woman, adman, adwoman.  
Gender-inclusive: ad-person, adperson, advertising executive, ad professional.  

\noindent \textless START\_Feminine\textgreater  
1. The ad-woman said she reviewed her presentation carefully before sending it to her manager.  
   Gender indicators: [ad-woman, she, her]  
2. An adwoman explained that she trusted her creative instincts.  
   Gender indicators: [adwoman, she, her]  
\textless END\_Feminine\textgreater  

\textless START\_Masculine\textgreater  
1. The ad-man said he revised his campaign proposal.  
   Gender indicators: [ad-man, he, his]  
2. An adman noted that he admired his colleague’s honesty.  
   Gender indicators: [adman, he, his]  
\textless END\_Masculine\textgreater  

\textless START\_Gender-neutral\textgreater  
1. The ad-person said they updated their project notes before the deadline.  
   Gender indicators: [ad-person, they, their]  
2. An advertising executive explained that they trusted their judgment.  
   Gender indicators: [advertising executive, they, their]  
\textless END\_Gender-neutral\textgreater
\end{tcolorbox}
\caption{\InclusiveGender Prompt}\label{fig:inclusivegender-prompt}
\end{figure*}

\clearpage

\section{Layer-wise Gender Neuron Distribution} \label{appx:neuron-distribution}

We analyze the distribution of identified gender neurons across model layers to understand where gender-related information is encoded. \texttt{Llama 3.1} consists of 32 transformer layers, while \texttt{Qwen 2.5} has 28 layers. We group layers into four ranges for comparison: early layers (0--5), early-middle layers (6--15), late-middle layers (16--25), and final layers (26+).

Table~\ref{tab:appx_neuron_distribution} shows the percentage of identified neurons in each layer group across both datasets. Our method identifies neurons that are strongly concentrated in the early layers (0--5), with 47.3\% and 41.1\% for \texttt{Llama 3.1}, and 78.4\% and 90.8\% for \texttt{Qwen 2.5} on \InclusiveGender and \cggender respectively. 

In contrast, sNeuron-TST selects a fixed number of neurons per layer without considering layer-specific activation patterns, resulting in uniform distribution across all layer groups. LAPE shows a more scattered distribution, with notable presence in middle and later layers. For \texttt{Qwen 2.5} on \cggender, LAPE identifies 41.9\% of neurons in layers 16--25, suggesting different selection criteria compared to our approach.

The concentration of our identified neurons in early layers indicates that gender information is primarily processed in the initial stages of the model. This observation has practical implications: interventions targeting these early-layer neurons may be more effective for controlling gender-related outputs while minimizing disruption to other model capabilities.
\begin{table*}[ht]
\centering
\footnotesize
\resizebox{0.9\textwidth}{!}{%
\begin{tabular*}{\textwidth}{@{\extracolsep\fill}llcccc}
\toprule
\textbf{Model} & \textbf{Method} &
\textbf{Layers 0--5} &
\textbf{Layers 6--15} &
\textbf{Layers 16--25} &
\textbf{Layers 26+} \\
\midrule
\multicolumn{6}{l}{\InclusiveGender} \\
\midrule
Llama 3.1 & Ours & 47.3\% & 23.1\% & 16.8\% & 12.8\% \\
          & sNeuron-TST & 33.3\% & 33.3\% & 16.7\% & 16.7\% \\
          & LAPE & 35.2\% & 31.4\% & 19.6\% & 13.8\% \\
\midrule
Qwen 2.5 & Ours & 78.4\% & 8.7\% & 2.4\% & 10.5\% \\
         & sNeuron-TST & 33.3\% & 33.3\% & 16.7\% & 16.7\% \\
         & LAPE & 37.8\% & 28.6\% & 21.3\% & 12.3\% \\
\midrule
\multicolumn{6}{l}{\cggender} \\
\midrule
Llama 3.1 & Ours & 41.1\% & 5.5\% & 11.0\% & 42.5\% \\
          & sNeuron-TST & 33.3\% & 33.3\% & 16.7\% & 16.7\% \\
          & LAPE & 40.4\% & 27.1\% & 14.6\% & 17.9\% \\
\midrule
Qwen 2.5 & Ours & 90.8\% & 3.0\% & 1.9\% & 4.3\% \\
         & sNeuron-TST & 33.3\% & 33.3\% & 16.7\% & 16.7\% \\
         & LAPE & 32.4\% & 22.8\% & 41.9\% & 2.8\% \\
\bottomrule
\end{tabular*}
}
\caption{Distribution of identified neurons across model layers on \InclusiveGender and \cggender. Values indicate the percentage of total identified neurons per layer group.}
\label{tab:appx_neuron_distribution}
\end{table*}

\section{Human Annotation Instruction} \label{appx:human_instruction}
\begin{figure}[ht]
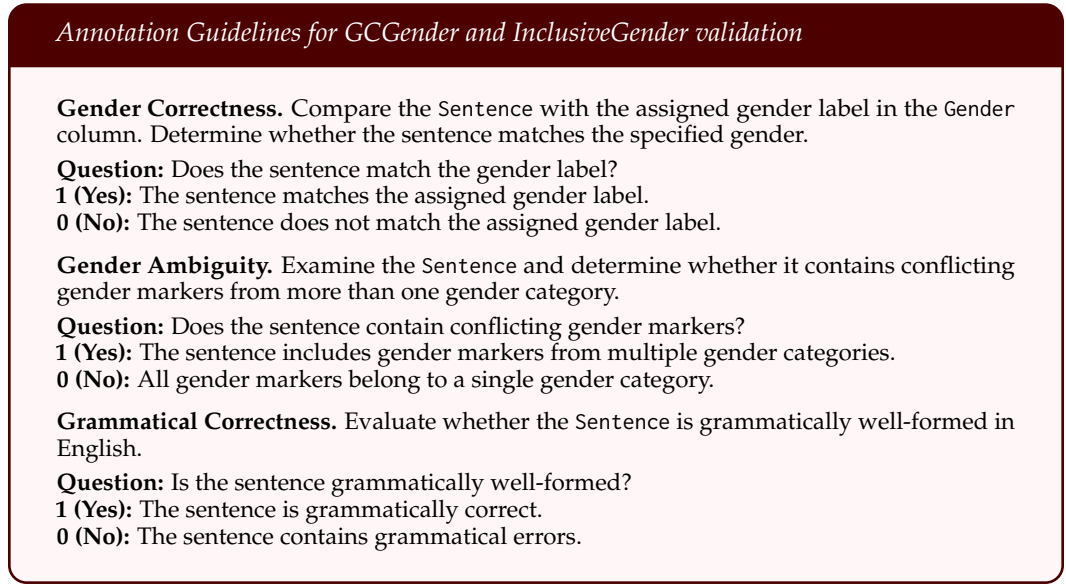

\begin{tcolorbox}[
  enhanced,
  breakable,
  title=\textit{Annotation Guidelines for \cggender and \InclusiveGender validation},
  colframe=red!30!black,
  colback=red!3!white,
  coltitle=white,
  colbacktitle=red!30!black,
  boxrule=1pt,
  boxsep=6pt,
  arc=6pt
]
\footnotesize
\noindent\textbf{Gender Correctness.} Compare the \texttt{Sentence} with the assigned gender label in the \texttt{Gender} column. Determine whether the sentence matches the specified gender.
\smallskip

\textbf{Question:} Does the sentence match the gender label?\\
\textbf{1 (Yes):} The sentence matches the assigned gender label.\\
\textbf{0 (No):} The sentence does not match the assigned gender label.
\medskip

\noindent\textbf{Gender Ambiguity.} Examine the \texttt{Sentence} and determine whether it contains conflicting gender markers from more than one gender category.
\smallskip

\textbf{Question:} Does the sentence contain conflicting gender markers?\\
\textbf{1 (Yes):} The sentence includes gender markers from multiple gender categories.\\
\textbf{0 (No):} All gender markers belong to a single gender category.
\medskip

\noindent\textbf{Grammatical Correctness.} Evaluate whether the \texttt{Sentence} is grammatically well-formed in English.
\smallskip

\textbf{Question:} Is the sentence grammatically well-formed?\\
\textbf{1 (Yes):} The sentence is grammatically correct.\\
\textbf{0 (No):} The sentence contains grammatical errors.
\end{tcolorbox}
\caption{Annotation guidelines for validating the \cggender and \InclusiveGender datasets. 
Each sentence is assessed along three binary dimensions: whether it matches its assigned gender label, whether it contains conflicting gender markers, and whether it is grammatically well-formed.
}

\label{fig:human-instruction-data}
\end{figure}

\begin{figure}[ht]
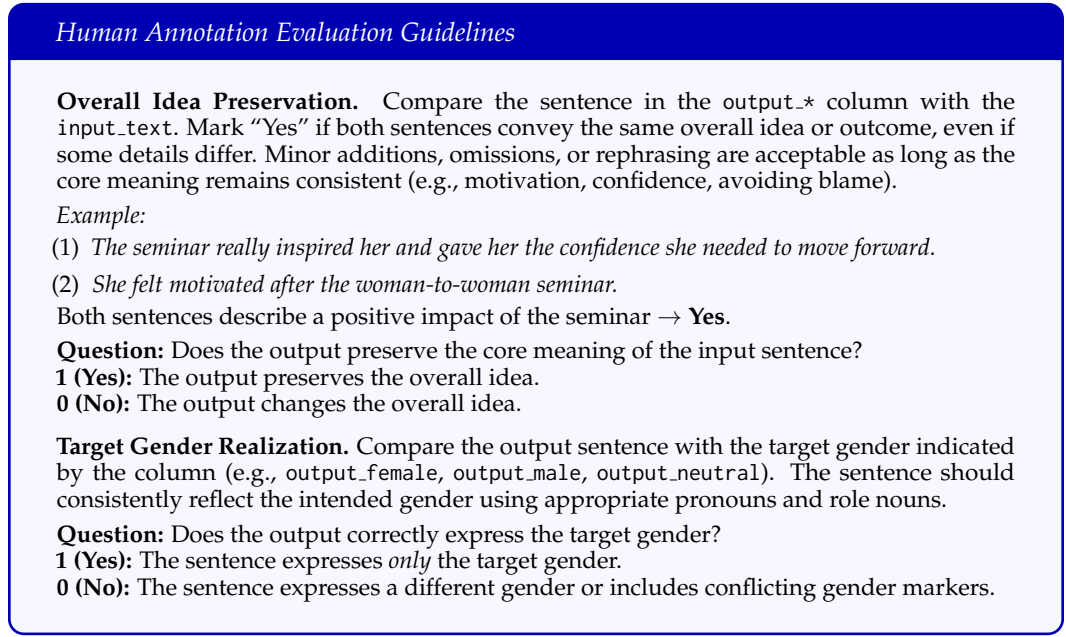

\begin{tcolorbox}[
  enhanced,
  breakable,
  colframe=blue!70!black,
  colback=blue!3!white,
  coltitle=white,
  boxrule=1pt,
  boxsep=6pt,
  arc=5pt,
  title=\textit{Human Annotation Evaluation Guidelines}
]
\footnotesize

\noindent\textbf{Overall Idea Preservation.} Compare the sentence in the \texttt{output\_*} column with the \texttt{input\_text}. Mark ``Yes'' if both sentences convey the same overall idea or outcome, even if some details differ. Minor additions, omissions, or rephrasing are acceptable as long as the core meaning remains consistent (e.g., motivation, confidence, avoiding blame).

\smallskip
\textit{Example:}
\begin{itemize}[leftmargin=1.5em, itemsep=2pt, topsep=2pt]
  \item[(1)] \textit{The seminar really inspired her and gave her the confidence she needed to move forward.}
  \item[(2)] \textit{She felt motivated after the woman-to-woman seminar.}
\end{itemize}
Both sentences describe a positive impact of the seminar $\rightarrow$ \textbf{Yes}.

\smallskip
\textbf{Question:} Does the output preserve the core meaning of the input sentence?\\
\textbf{1 (Yes):} The output preserves the overall idea.\\
\textbf{0 (No):} The output changes the overall idea.

\medskip

\noindent\textbf{Target Gender Realization.} Compare the output sentence with the target gender indicated by the column (e.g., \texttt{output\_female}, \texttt{output\_male}, \texttt{output\_neutral}). The sentence should consistently reflect the intended gender using appropriate pronouns and role nouns.

\smallskip
\textbf{Question:} Does the output correctly express the target gender?\\
\textbf{1 (Yes):} The sentence expresses \emph{only} the target gender.\\
\textbf{0 (No):} The sentence expresses a different gender or includes conflicting gender markers.

\end{tcolorbox}
\caption{Annotation guidelines for human evaluation of gender transformation quality.
Annotators assess each output along two dimensions: whether the core meaning of the input is preserved, and whether the output consistently reflects the target gender.
}

\label{fig:human-instruction-eval}
\end{figure}

We worked with human annotators to evaluate both the quality of our synthetic datasets and the quality of gender transformation outputs. We recruited two graduate students with strong English proficiency and familiarity with linguistic analysis to perform the annotation tasks. To ensure consistency and alignment with the evaluation criteria, annotators participated in a brief training session and reviewed example annotations provided by the authors. Annotators were compensated based on hours worked. For the annotation of gender transformation outputs, annotators were not informed about the source of the outputs (e.g., baseline vs.\ our method) to avoid potential bias. Annotations were performed independently by two annotators using predefined guidelines (see figures~\ref{fig:human-instruction-data} and~\ref{fig:human-instruction-eval}).
\clearpage

\clearpage

\section{Prompt Templates} \label{appx:prompts}
\begin{figure}[ht]
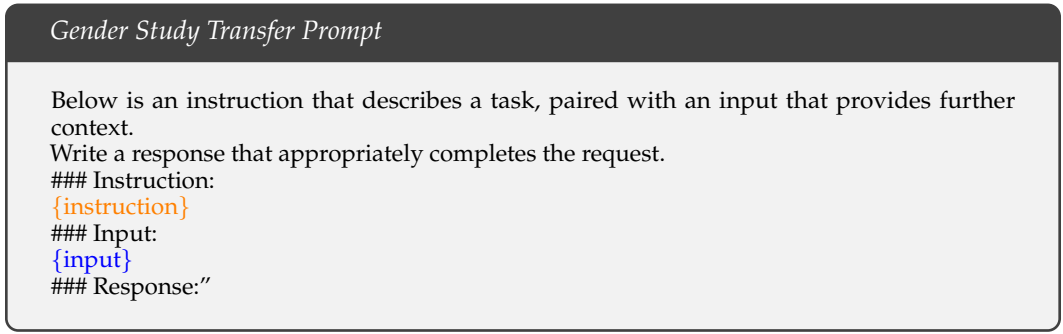

\begin{tcolorbox}[
  enhanced,
  breakable,
  boxrule=1pt,
  boxsep=5pt,
  arc=4pt,
  title=\textit{Gender Study Transfer Prompt}
]
\footnotesize
\noindent Below is an instruction that describes a task, paired with an input that provides further context. 
\\
Write a response that appropriately completes the request.

\#\#\# Instruction:\\
\textcolor{orange}{\{instruction\}}

\#\#\# Input:\\
\textcolor{blue}{\{input\}}

\#\#\# Response:"
\end{tcolorbox}
\caption{Gender Study Transfer Prompt}\label{fig:gender-transfer-prompt}
\end{figure}

\begin{figure}[ht]
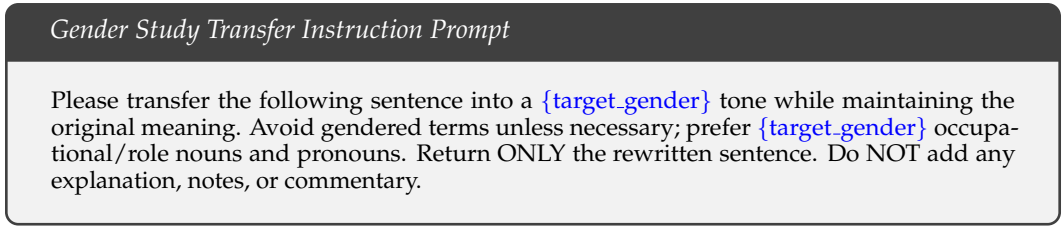

\begin{tcolorbox}[
  enhanced,
  breakable,
  boxrule=1pt,
  boxsep=5pt,
  arc=4pt,
  title=\textit{Gender Study Transfer Instruction Prompt}
]
\footnotesize
\noindent Please transfer the following sentence into a \textcolor{blue}{\{target\_gender\}} tone while maintaining the original meaning. Avoid gendered terms unless necessary; prefer \textcolor{blue}{\{target\_gender\}} occupational/role nouns and pronouns. Return ONLY the rewritten sentence. Do NOT add any explanation, notes, or commentary.
\end{tcolorbox}
\caption{Gender Study Transfer Instruction Prompt}\label{fig:gender-transferinstruction-prompt}
\end{figure}

We report the gender transfer prompts used in the experiments (Tables~\ref{tab:merged_comparison},~\ref{tab:qwen_comparison}, and~\ref{tab:llama_comparison}). Specifically, we prompt the LMs to transform the given input sentence into a specific gender category with or without gender neuron masking (Figures \ref{fig:gender-transfer-prompt}-\ref{fig:gender-transferinstruction-prompt}). 

\section{Gender Neuron Identification Full Table Results}\label{appx:full-tables}
Tables~\ref{tab:qwen_comparison} and~\ref{tab:llama_comparison} report the full gender neuron masking results for \texttt{Qwen 2.5} and \texttt{Llama 3.1}, respectively. Each table reports both the average number of gendered terms per response and the change in mention ratios ($\Delta$ Ratio) relative to the unmasked baseline under the same target. The main paper (Table~\ref{tab:merged_comparison}) summarises the $\Delta$ Ratio columns only.

\begin{table*}[ht]
\centering
\footnotesize
\resizebox{0.9\textwidth}{!}{%
\begin{tabular}{l|ccc|ccc}
\toprule
\multirow{2}{*}{\textbf{Method}} &
\multicolumn{3}{c|}{\textbf{Num. of Gendered Terms per Response}} &
\multicolumn{3}{c}{\textbf{$\Delta$ Ratio (\%)}} \\
\cmidrule(lr){2-4}\cmidrule(lr){5-7}
& M & F & N & M & F & N \\
\midrule
\multicolumn{7}{c}{\textbf{\InclusiveGender — Feminine}} \\
\midrule
Baseline & 0.07 (0.26\%) & 2.10 (7.98\%) & 0.34 (1.29\%) & — & — & — \\
sNeuron-TST & 0.10 (0.18\%) & 2.27 (4.18\%) & 1.86 (3.43\%) & -0.09 & -3.80 & +2.14 \\
LAPE & 0.14 (0.33\%) & 1.75 (3.98\%) & 0.67 (1.52\%) & +0.07 & -4.00 & +0.23 \\
Ours & 0.06 (0.22\%) & 2.18 (8.15\%) & 0.32 (1.20\%) & \cellcolor[HTML]{D5F5E3}\textbf{-0.04} & \cellcolor[HTML]{D5F5E3}\textbf{+0.17} & \cellcolor[HTML]{D5F5E3}\textbf{-0.09} \\
\midrule
\multicolumn{7}{c}{\textbf{\InclusiveGender — Masculine}} \\
\midrule
Baseline & 1.70 (5.91\%) & 0.05 (0.16\%) & 0.41 (1.43\%) & — & — & — \\
sNeuron-TST & 1.87 (6.08\%) & 0.07 (0.23\%) & 1.12 (3.65\%) & +0.17 & +0.07 & +2.23 \\
LAPE & 1.69 (3.88\%) & 0.02 (0.05\%) & 0.73 (1.69\%) & -2.03 & -0.11 & +0.26 \\
Ours & 1.70 (5.93\%) & 0.04 (0.13\%) & 0.40 (1.31\%) & \cellcolor[HTML]{D5F5E3}\textbf{+0.02} & \cellcolor[HTML]{D5F5E3}\textbf{-0.03} & \cellcolor[HTML]{D5F5E3}\textbf{-0.03} \\
\midrule
\multicolumn{7}{c}{\textbf{\InclusiveGender — Neutral}} \\
\midrule
Baseline & 0.14 (0.95\%) & 0.03 (0.23\%) & 2.08 (14.19\%) & — & — & — \\
sNeuron-TST & 0.21 (2.07\%) & 0.06 (0.53\%) & 1.82 (17.60\%) & +1.11 & +0.30 & +3.40 \\
LAPE & 0.09 (0.79\%) & 0.03 (0.29\%) & 2.15 (20.00\%) & -0.16 & +0.06 & +5.81 \\
Ours & 0.13 (0.92\%) & 0.03 (0.22\%) & 2.08 (14.42\%) & \cellcolor[HTML]{D5F5E3}\textbf{-0.03} & \cellcolor[HTML]{D5F5E3}\textbf{-0.01 }& \cellcolor[HTML]{D5F5E3}\textbf{+0.22} \\
\midrule
\multicolumn{7}{c}{\textbf{\cggender — Feminine}} \\
\midrule
Baseline & 0.16 (0.54\%) & 1.19 (4.06\%) & 0.46 (1.57\%) & — & — & — \\
sNeuron-TST & 0.20 (0.39\%) & 1.14 (2.24\%) & 0.43 (0.85\%) & +0.02 & -1.53 & -0.34 \\
LAPE & 0.15 (0.31\%) & 0.96 (1.95\%) & 0.65 (1.32\%) & -0.06 & -1.83 & +0.13 \\
Ours & 0.16 (0.53\%) & 1.22 (4.12\%) & 0.45 (1.52\%) & \cellcolor[HTML]{D5F5E3}\textbf{-0.01} & \cellcolor[HTML]{D5F5E3}\textbf{+0.07} & \cellcolor[HTML]{D5F5E3}\textbf{-0.05} \\
\midrule
\multicolumn{7}{c}{\textbf{\cggender — Masculine}} \\
\midrule
Baseline & 0.89 (2.51\%) & 0.21 (0.60\%) & 0.47 (1.32\%) & — & — & — \\
sNeuron-TST & 0.86 (1.67\%) & 0.03 (0.06\%) & 2.41 (4.71\%) & -0.83 & -0.54 & +3.39 \\
LAPE & 0.95 (1.94\%) & 0.11 (0.22\%) & 0.57 (1.16\%) & -0.57 & -0.38 & -0.16 \\
Ours & 0.83 (2.69\%) & 0.22 (0.72\%) & 0.46 (1.49\%) & \cellcolor[HTML]{D5F5E3}\textbf{+0.06} & \cellcolor[HTML]{D5F5E3}\textbf{-0.11} & \cellcolor[HTML]{D5F5E3}\textbf{-0.02} \\
\midrule
\multicolumn{7}{c}{\textbf{\cggender — Neutral}} \\
\midrule
Baseline & 0.03 (0.10\%) & 0.02 (0.09\%) & 1.65 (6.43\%) & — & — & — \\
sNeuron-TST & 0.06 (0.36\%) & 0.01 (0.05\%) & 1.61 (8.99\%) & +0.24 & -0.04 & +1.61 \\
LAPE & 0.03 (0.12\%) & 0.03 (0.11\%) & 1.69 (6.33\%) & -0.00 & +0.02 & -1.05 \\
Ours & 0.03 (0.08\%) & 0.01 (0.06\%) & 1.65 (6.42\%) & \cellcolor[HTML]{FFF9C4}\textbf{-0.02} & \cellcolor[HTML]{FFF9C4}\textbf{-0.04} & \cellcolor[HTML]{FFF9C4}\textbf{-0.01} \\

\bottomrule
\end{tabular}%
}
\caption{Gender neuron masking results for \texttt{Qwen 2.5} on \InclusiveGender and \cggender. Masked columns show terms per response (ratio), $\Delta$ Ratio shows the change from baseline. Num.\ of Gendered Terms per Response reports the average count of matched lexicon terms for Masculine (M), Feminine (F), and Neutral (N). Values in parentheses are the corresponding mention ratios of gendered terms (\% of all generated words). $\Delta$ Ratio (\%) is the change in gendered term mention ratio (percentage points) compared to the Baseline row in the same block (positive = increase, negative = decrease). For our method, \smash{\setlength{\fboxsep}{1.2pt}\colorbox[HTML]{D5F5E3}{green}} indicates rows where our method achieves the best target-consistent control, while \smash{\setlength{\fboxsep}{1.2pt}\colorbox[HTML]{FFF9C4}{yellow}} marks rows where no method fully succeeds but ours shows the least leakage.}
\label{tab:qwen_comparison}
\end{table*}
\begin{table*}[ht]
\centering
\footnotesize
\resizebox{0.9\textwidth}{!}{%
\begin{tabular}{l|ccc|ccc}
\toprule
\multirow{2}{*}{\textbf{Method}} &
\multicolumn{3}{c|}{\textbf{Num. of Gendered Terms per Response}} &
\multicolumn{3}{c}{\textbf{$\Delta$ Ratio (\%)}} \\
\cmidrule(lr){2-4}\cmidrule(lr){5-7}
& M & F & N & M & F & N \\
\midrule
\multicolumn{7}{c}{\textbf{\InclusiveGender — Feminine}} \\
\midrule
Baseline & 0.38 (2.60\%) & 1.10 (7.55\%) & 0.33 (2.23\%) & — & — & — \\
sNeuron-TST & 0.80 (6.14\%) & 0.66 (3.18\%) & 0.32 (2.21\%) & +3.54 & -4.37 & -0.02 \\
LAPE & 0.28 (2.24\%) & 1.03 (7.36\%) & 0.35 (2.27\%) & -0.36 & -0.19 & +0.04 \\
Ours & 0.33 (2.27\%) & 1.22 (8.44\%) & 0.31 (2.19\%) & \cellcolor[HTML]{D5F5E3}\textbf{-0.33} & \cellcolor[HTML]{D5F5E3}\textbf{+0.89} & \cellcolor[HTML]{D5F5E3}\textbf{-0.04} \\
\midrule
\multicolumn{7}{c}{\textbf{\InclusiveGender — Masculine}} \\
\midrule
Baseline & 1.58 (13.22\%) & 0.11 (0.89\%) & 0.20 (1.68\%) & — & — & — \\
sNeuron-TST & 1.56 (10.85\%) & 0.40 (2.76\%) & 0.25 (1.76\%) & -2.37 & +1.87 & +0.08 \\
LAPE& 1.39 (11.57\%) & 0.18 (1.48\%) & 0.25 (2.10\%) & -1.65 & +0.59 & +0.42 \\
Ours & 1.60 (13.55\%) & 0.19 (1.54\%) & 0.17 (1.58\%) & \cellcolor[HTML]{D5F5E3}\textbf{+0.33} & \cellcolor[HTML]{D5F5E3}\textbf{-0.65} & \cellcolor[HTML]{D5F5E3}\textbf{-0.10} \\
\midrule
\multicolumn{7}{c}{\textbf{\InclusiveGender — Neutral}} \\
\midrule
Baseline & 0.04 (0.43\%) & 0.01 (0.07\%) & 2.35 (22.44\%) & — & — & — \\
sNeuron-TST & 0.28 (0.60\%) & 0.15 (0.32\%) & 1.59 (12.02\%) & +0.17 & +0.25 & -10.42 \\
LAPE& 0.06 (0.45\%) & 0.02 (0.08\%) & 2.30 (21.81\%) & +0.02 & +0.01 & -0.63 \\
Ours & 0.03 (0.41\%) & 0.01 (0.07\%) & 2.33 (22.13\%) & \cellcolor[HTML]{FFF9C4}\textbf{-0.02} & \cellcolor[HTML]{FFF9C4}\textbf{+0.00} & \cellcolor[HTML]{FFF9C4}\textbf{-0.31} \\
\midrule
\multicolumn{7}{c}{\textbf{\cggender — Feminine}} \\
\midrule
Baseline & 0.23 (1.24\%) & 1.29 (6.91\%) & 0.39 (2.10\%) & — & — & — \\
sNeuron-TST & 0.71 (3.69\%) & 0.53 (2.76\%) & 0.49 (2.55\%) & +2.45 & -4.14 & +0.45 \\
LAPE& 0.24 (1.22\%) & 1.26 (6.58\%) & 0.43 (2.25\%) & -0.02 & -0.33 & +0.15 \\
Ours & 0.22 (1.22\%) & 1.37 (7.01\%) & 0.38 (2.09\%) & \cellcolor[HTML]{D5F5E3}\textbf{-0.02} & \cellcolor[HTML]{D5F5E3}\textbf{+0.10} & \cellcolor[HTML]{D5F5E3}\textbf{-0.01} \\
\midrule
\multicolumn{7}{c}{\textbf{\cggender — Masculine}} \\
\midrule
Baseline & 1.08 (5.86\%) & 0.22 (1.18\%) & 0.31 (1.71\%) & — & — & — \\
sNeuron-TST & 1.01 (4.94\%) & 0.19 (0.91\%) & 0.28 (1.37\%) & -0.92 & -0.27 & -0.34 \\
LAPE& 1.00 (5.24\%) & 0.22 (1.15\%) & 0.34 (1.78\%) & -0.62 & -0.03 & +0.07 \\
Ours & 1.21 (6.62\%) & 0.15 (0.80\%) & 0.30 (1.69\%) & \cellcolor[HTML]{D5F5E3}\textbf{+0.76} & \cellcolor[HTML]{D5F5E3}\textbf{-0.38} & \cellcolor[HTML]{D5F5E3}\textbf{-0.02} \\
\midrule
\multicolumn{7}{c}{\textbf{\cggender — Neutral}} \\
\midrule
Baseline & 0.00 (0.01\%) & 0.01 (0.06\%) & 1.69 (10.12\%) & — & — & — \\
sNeuron-TST & 0.03 (0.06\%) & 0.01 (0.02\%) & 4.09 (8.97\%) & +0.04 & -0.04 & -1.16 \\
LAPE& 0.01 (0.04\%) & 0.00 (0.03\%) & 1.84 (9.66\%) & +0.02 & -0.03 & -0.46 \\
Ours & 0.00 (0.01\%) & 0.01 (0.06\%) & 1.68 (10.13\%) & \cellcolor[HTML]{FFF9C4}\textbf{+0.00} & \cellcolor[HTML]{FFF9C4}\textbf{+0.00} & \cellcolor[HTML]{FFF9C4}\textbf{+0.00} \\
\bottomrule
\end{tabular}%
}
\caption{Gender neuron masking results for \texttt{Llama 3.1} on \InclusiveGender and \cggender. Masked columns show terms per response (ratio), $\Delta$ Ratio shows the change from baseline. Num.\ of Gendered Terms per Response reports the average count of matched lexicon terms for Masculine (M), Feminine (F), and Neutral (N). Values in parentheses are the corresponding mention ratios of gendered terms (\% of all generated words). $\Delta$ Ratio (\%) is the change in gendered term mention ratio (percentage points) compared to the Baseline row in the same block (positive = increase, negative = decrease). For our method, \smash{\setlength{\fboxsep}{1.2pt}\colorbox[HTML]{D5F5E3}{green}} indicates rows where our method achieves the best target-consistent control, while \smash{\setlength{\fboxsep}{1.2pt}\colorbox[HTML]{FFF9C4}{yellow}} marks rows where no method fully succeeds but ours shows the least leakage.}
\label{tab:llama_comparison}
\end{table*}

\clearpage
\section{Case Study} \label{appx:case-study}
\begin{table*}[ht]
\centering
\scriptsize
\begin{tabular}{p{2.2cm}p{2.5cm}p{8cm}}
\toprule
& \multicolumn{1}{l}{\textbf{Prompt:}} & Please transfer the following sentence into a \{target\_gender\} tone while maintaining the original meaning. Avoid gendered terms unless necessary; prefer \{target\_gender\} occupational/role nouns and pronouns. Return ONLY the rewritten sentence. Do NOT add any explanation, notes, or commentary. \\ 
\midrule
\rowcolor{gray!20}
Neutral  & \textbf{Source Text:} & The police officer said they would handle their responsibilities themself. \\
\rowcolor{gray!20}
$\downarrow$ & \textbf{Baseline:} & The police said she would take care of her duties. \\
\rowcolor{gray!20}
Feminine & \textbf{Ours:} & The policewoman said she would handle her duties herself. \\
\midrule
Neutral & \textbf{Source Text:} & The police officer said they would handle their responsibilities themself. \\
$\downarrow$ & \textbf{Baseline:} & The cop said said he'd take care of his duties himself.\ \\
Masculine & \textbf{Ours:} & The policeman said he would take care of his duties himself. \\
\midrule
\rowcolor{gray!20}
Masculine & \textbf{Source Text:} & The fireman rescued the child and carried him to safety.\\
\rowcolor{gray!20}
$\downarrow$ & \textbf{Baseline:} & The firewoman rescued the child and carried her to safety. \\
\rowcolor{gray!20}
Feminine & \textbf{Ours:} & The woman firefighter rescued the girl and carried her out of danger \\
\midrule
Masculine& \textbf{Source Text:} & The fireman rescued the child and carried him to safety. \\
$\downarrow$ & \textbf{Baseline:} & The firefighter rescued the child and carried them to safety. \\
Neutral & \textbf{Ours:} & The firefighter rescued the child escape and ensured their safety. \\
\midrule
\rowcolor{gray!20}
Feminine & \textbf{Source Text:} & The chairwoman announced her decision to the board members. \\
\rowcolor{gray!20}
$\downarrow$ & \textbf{Baseline:} & The chairman announced the decision to the board members. \\
\rowcolor{gray!20}
Masculine & \textbf{Ours:} & The man communicated his decision to the board. \\
\midrule
Feminine & \textbf{Source Text:} & The chairwoman announced her decision to the board members. \\
$\downarrow$ & \textbf{Baseline:} & The chair announced the decision to the board members. \\
Neutral & \textbf{Ours:} & The chairperson announced their decision to the board members. \\
\bottomrule
\end{tabular}
\caption{Case study on gender transfer tasks. For each transfer, we keep only the neurons identified for the target gender active.}
\label{tab:gender_case_study}
\end{table*}

As shown in Table~\ref{tab:gender_case_study}, we present qualitative examples of gender transfer using our neuron masking approach. We evaluate six transfer directions across three gender categories: feminine, masculine, and gender-neutral.
The baseline setting may produce incomplete or inconsistent transformations. For instance, in the Neutral→Feminine transfer, the baseline generates ``The police said she would take care of her duties,'' which updates pronouns but retains the gender-neutral occupation term. In contrast, our method produces ``The policewoman said she would handle her duties herself,'' achieving a complete transformation of both the occupational noun and pronouns. Similarly, for Feminine→Neutral transfer, the baseline outputs ``The chair announced the decision,'' which removes gendered language but also drops the pronoun entirely. Our method generates ``The chairperson announced their decision to the board members,'' correctly substituting singular they/their to maintain grammatical completeness while achieving neutrality.

These examples demonstrate that selectively activating target-gender neurons enables more comprehensive gender transfer. Compared to the baseline, our targeted gender neuron masking produces more consistent gender transformations with less non-target leakage.

\end{document}